\documentclass{article}

\usepackage[final]{corl_2023} 

\usepackage{xspace}
\usepackage[normalem]{ulem}
\newcommand{\appr}{PLEX\xspace}
\newcommand{\mtvd}{\mathcal{D}_{\textrm{mtvd}}\xspace}
\newcommand{\vmt}{\mathcal{D}_{\textrm{vmt}}\xspace}
\newcommand{\ttd}{\mathcal{D}_{\textrm{ttd}}\xspace}
\newcommand{\mw}{Meta-World\xspace}
\newcommand{\rsrw}{Robosuite\xspace}
\usepackage{wrapfig}
\usepackage[symbol]{footmisc}

\usepackage{amsmath}
\usepackage{amsfonts}
\usepackage{amssymb}
\usepackage{amsthm}
\usepackage{bm}
\usepackage{bbm}
\usepackage{mathtools}
\usepackage[inline]{enumitem}
\usepackage{mathrsfs} 
\usepackage{wrapfig}
\usepackage{thmtools,thm-restate}
\usepackage{algorithm}
\usepackage{algorithmic}
\usepackage[capitalise,noabbrev]{cleveref}
\usepackage{color}
\usepackage{xcolor}
\usepackage{graphicx}
\usepackage{makecell}
\usepackage{comment}
\usepackage[page]{appendix}

\theoremstyle{plain}

\theoremstyle{definition}

\theoremstyle{remark}

\def\AA{\mathcal{A}}

\def\GG{\mathcal{G}}
\def\LL{\mathcal{L}}
\def\OO{\mathcal{O}}
\def\PP{\mathcal{P}}
\def\SS{\mathcal{S}}
\def\WW{\mathcal{W}}

\def\*{\star}
\DeclareMathSymbol{\mhef}{\mathord}{operators}{`\-}

\newcommand{\replaced}[2]{#2}

\usepackage{graphicx}

\title{\appr: Making the Most of the Available Data\\ for Robotic Manipulation Pretraining}

\author{
  Garrett Thomas\thanks{Stanford University, 
  \texttt{gwthomas@stanford.edu}. Work done partly while at Microsoft Research.}\\
  \And
  Ching-An Cheng\thanks{Microsoft Research,
  \texttt{\{chinganc,riloynd,fevieira,vivineet,akolobov\}@microsoft.com}}\\
  \And
  Ricky Loynd\footnotemark[2]\\
  \And
  Felipe Vieira Frujeri\footnotemark[2]\\
  \AND
  Vibhav Vineet\footnotemark[2]\\
  \And
  Mihai Jalobeanu\thanks{dexman.ai, 
  \texttt{mihai@dexman.ai}. Work done partly while at Microsoft Research.}\\
  \And
  Andrey Kolobov\footnotemark[2]
}

\begin{document}

\maketitle


\begin{abstract}
A rich representation is key to general robotic manipulation, but existing approaches to representation learning require large amounts of multimodal demonstrations.
In this work we propose \appr\footnote{The paper's accompanying code and data are available at \href{https://microsoft.github.io/PLEX}{https://microsoft.github.io/PLEX}.}, a transformer-based architecture that learns from a small amount of \emph{task-agnostic visuomotor} trajectories and a much larger amount of \emph{task-conditioned} object manipulation \emph{videos} --- a type of data available in quantity. 
\appr\ uses visuomotor trajectories to induce a latent feature space and to learn task-agnostic manipulation routines, while diverse video-only demonstrations teach \appr\ how to plan in the induced latent feature space for a wide variety of tasks. Experiments showcase \appr's generalization on \mw\ and SOTA performance in challenging \rsrw\ environments. In particular, using relative positional encoding in \appr's transformers greatly helps in low-data regimes of learning from human-collected demonstrations.
\end{abstract}



\section{Introduction}
\vspace{-1mm}

Transformers~\cite{vaswani2017attention} have lead to breakthroughs in training large-scale general representations for computer vision (CV) and natural language processing (NLP)~\cite{brown2020gpt3}, enabling zero-shot adaptation and fast finetuning~\citep{fm2021}. At the same time, despite impressive progress, transformer-based representations haven't shown the same versatility for robotic manipulation. Some attribute this gap to the lack of suitable training data for robotics~\citep{fm2021}. We argue instead that data relevant to training robotic manipulation models is copious but has important structure that most existing training methods ignore and fail to leverage. These insights lead us to propose a novel transformer-based  \replaced{}{architecture}, called \emph{\appr}, that is capable of \replaced{zero-shot adaptation and effective finetuning thanks to being tailored to the realities of robotic manipulation data}{effective learning from realistically available robotic manipulation datasets}. 

\replaced{We observe that robotics-relevant data falls into three major categories. \textbf{(1)} The most plentiful category is comprised of ``in-the-wild'' video datasets~\cite{youtue8m}~\cite{ho2022imagenvideo}~\cite{grauman2022ego4d}. Some of them, e.g., Epic Kitchens~\cite{Damen2018EPICKITCHENS}~\cite{Damen2022RESCALING}, focus on object manipulation. In aggregation, these datasets cover an immense variety of tasks and are typically annotated with activity descriptions but contain no explicit action information for a robotic to mimic. \textbf{(2)} The second category consists of matching sequences of percepts and actions. In some datasets of this type, these sequences don't correspond to meaningful task and are generated by a scripted exploration policy~\cite{robonetv1}. In others, they come from well-defined tasks, but even in the largest such datasets, e.g.,~\citep{robonetv2}, the task coverage is modest compared to video-only data such as~\cite{grauman2022ego4d}. Nonetheless, these sensorimotor sequences capture valuable correlations between a robot's actions and changes in the environment. \textbf{(3)} The third data category, the scarcest one, consists of high-quality demonstrations for a target task in a target environment. Thus, a scalable model architecture for robotic manipulation must be able to learn {primarily} from videos, while being extra data-efficient on sensorimotor training sequences and the available demonstrations from the target environment.}{We observe that robotics-relevant data falls into three major categories: \textbf{(1)} Video-only data, which contain high-quality and potentially description-annotated demonstrations for an immense variety of tasks but have no explicit action information for a robot to mimic; \textbf{(2)} Data containing matching sequences of percepts \emph{and actions}, which are less plentiful than pure videos and don't necessarily correspond to meaningful tasks~\cite{robonetv1}, but capture valuable correlations between a robot's actions and changes in the environment and are easy to collect on a given robot; \textbf{(3)} Small sets of high-quality sensorimotor demonstrations for a target task in a target environment. Thus, a scalable model architecture for robotic manipulation must be able to learn {primarily} from videos, while being extra data-efficient on sensorimotor training sequences and the small amount target demonstrations.}

\appr, the \textit{\textbf{PL}}anning-\textit{\textbf{EX}}ecution architecture we propose, is designed to take advantage of data sources of these types. A \appr\ model has two major transformer-based components: \textbf{(I)} a task-conditioned {observational} \emph{planner} that, given a task specification and an estimate of the current world state, determines the next state to which the robot should attempt to transition, and \textbf{(II)} an \emph{executor} that, having received the desired next state from the planner, produces an action that should lead there from the current state. The executor is trained by optimizing an inverse dynamics loss over exploratory sensorimotor data of the aforementioned category \textbf{(2)}, while the planner is trained by minimizing a loss of its autoregressive predictions computed with respect to video-only trajectories of category \textbf{(1)}. The target-task data of category \textbf{(3)} can be optionally used to efficiently finetune the planner, the executor, or both.

We make three design choices that greatly help the data efficiency of \appr's training:

\vspace{-1mm}
\begin{itemize}[leftmargin=*]    
\item \emph{Learning to plan in the observation embedding space.} Rather than generating videos of proposed task execution using, e.g., stable diffusion as in~\citet{du2023unipi}, \appr\ learns to plan and execute in the low-dimensional space of observation embeddings.

\item\emph{Asymmetric learning of the embedding space.} The \replaced{}{observation embedding} space in which the executor and the planner operate is induced \replaced{either by a pretrained frozen feature-rich observation encoder, such as R3M~\cite{nair2022r3m}, or by the executor's training loss \emph{only}. Keeping the planner's gradients from entering observation encoders makes training significantly cheaper and removes the risk of latent space collapse.}{by training the observation encoder using the executor's loss \emph{only} (or even by employing a frozen feature-rich encoder such as R3M~\cite{nair2022r3m}). The planner's gradients don't affect the encoder, which reduces the cost of \appr\ training.}

\item \emph{Relative positional encodings.} 
We adopt the relative positional encodings \cite{orig} in \appr. We empirically show that in robotic manipulation the relative positional encodings significantly improve training efficiency from human-collected data compared with the \emph{absolute} positional encodings~\cite{vaswani2017attention} commonly used in the literature on transformers.
\end{itemize}
\vspace{-1mm}
Most approaches that use video-only demonstrations for pretraining in robotic manipulation produce purely visual representations (see, e.g., \cite{yen2020see,chen2021learning,nair2022r3m,Radosavovic2022RWRL}). The majority of algorithms that produce sensorimotor models need most or all of the video demonstrations to be accompanied by action sequences that generated the videos, a requirement that holds only for a small fraction available manipulation data
~\cite{mcil21rss,jang2021bc,mandi2021generalizable,gato,robonetv2,nasiriany2022learning,rt2022}. Few approaches have a dedicated trainable planning component; e.g., 
\cite{hakhamaneshi2021hierarchical,ren2021generalization,xihan2022skill,nasiriany2022learning,mees2022matters} plan in a skill space, which \appr\ can be modified to do as well. Conceptually, \appr\ falls under the paradigm of learning from observations (LfO), but existing LfO approaches don't have multitask zero-shot planning capability~\cite{nair2017knot,rados2021soil,pathak2018zs,vpt2022} or demostrate it only in low-dimensional environments across similar tasks~\cite{xu2022pgiorl}. Of the works that have used transformers for robotic manipulation~\cite{dasari2020transformers,kim2021transformer,mees2022matters,gato,rt2022}, only \citet{rt2022} have analyzed their data efficiency, and none have looked at positional embeddings as a way to improve it. Overall, the closest approach to \appr\ is the concurrently proposed UniPi~\cite{du2023unipi}. It also has counterparts of \appr's planner and executor, but its planner operates using  diffusion \emph{in the image space}~\cite{ho2022imagenvideo}, which is expensive both datawise and computationally, and may fail to model manipulation-relevant 3D object structure consistently~\cite{ho2022imagenvideo}. A more extensive discussion of prior work is provided in \Cref{app:rel_work}.

We experimentally show that \appr's planner-executor design can effectively exploit the structure of realistically available robotic manipulation data to achieve efficient learning. 
On the multi-task \mw~\cite{mw2019} benchmark, despite pretraining mostly on video data, \appr\ exhibits strong zero-shot performance on unseen tasks and can be further improved by finetuning on a small amount of video-only demonstrations. We empirically show on the challenging Robosuite/Robomimic~\cite{robosuite,robomimic2021} benchmark that, contrary to conclusions from NLP~\cite{orig}, the use of relative positional encodings significantly improves the data efficiency of \appr\ learning from human-collected demonstrations. 

\vspace{-3mm}
\section{Problem statement and relevant concepts}
\vspace{-1mm}

\subsection{Problem statement}
\vspace{-1mm}

We consider the problem of learning a generalist  task-conditioned policy for goal-directed object manipulation. Namely, we seek a policy that can control a robotic manipulator to successfully accomplish tasks that the robot may not have encountered during the policy training process; such a policy formally can be viewed a solution to a task-conditioned partially observable Markov decision process (POMDP) described in \Cref{app:pomdp}. In practice, learning a generalist policy that performs well on a broad distribution of tasks zero-shot is very challenging, as the coverage and amount of publicly available training data are limited. Therefore, in this work we consider a two-phased learning process: (1) pretraining, during which a generalist policy is trained, and (2) finetuning, during which this policy is adapted to a target task.

\vspace{-1mm}
\subsection{Data for training robotic manipulation models}
\vspace{-1mm}
 
We consider three broad groups of datasets relevant to training robotic manipulation systems:\footnote{Static image datasets, e.g., ImageNet, aren't treated by \appr\ in a special way and we don't discuss it here, but can be used to pretrain \appr's image encoders.}

\textbf{Multi-task video demonstrations ($\mtvd$).}
Being the most abundant category, it comprises data collections ranging from general YouTube videos to curated benchmarks such as Ego4D~\cite{grauman2022ego4d}, Epic Kitchens~\cite{Damen2018EPICKITCHENS,Damen2022RESCALING}, and YouTube-8M~\cite{youtue8m} showing \emph{an} agent -- either a robot or a person -- performing a meaningful object manipulation task with an end-effector. This data contains demonstration-quality sequences of video observations and descriptions of tasks they accomplish, but not the action sequences whose execution generated these videos.
 
\textbf{Visuomotor trajectories ($\vmt$).} These trajectories consist of paired sequences of observations and robots' actions. Although some of them may be high-quality demonstrations of specific tasks, e.g., as in the Bridge Dataset~\cite{robonetv2}, many of these trajectories are generated by activities that most people will not find meaningful, e.g., grabbing random objects in a tray, as in the RoboNet~\cite{robonetv1}. Since no strong quality, quantity, or task association requirements are imposed on $\vmt$\ data, it is relatively easy to collect for any target embodiment and environment.

\textbf{Target-task demonstrations ($\ttd$)}. 
This is the most scarce but also most desirable data category, since it encompasses high-quality trajectories for a specific task in question, ideally collected on the target embodiment (robot). Note, however that we don't require that these demonstrations be visuomotor. In fact, our experiments show that \appr\ needs only video demonstrations for finetuning to learn a high-quality policy for a target task.

\textbf{A key data assumption} we make in this work is that $|\ttd | \ll |\vmt| \ll |\mtvd|$.

\vspace{-1mm}
\subsection{Transformers and positional encodings}
\vspace{-1mm}
A transformer-based architecture consists of several specially structured \emph{self-attention layers} and, in general, maps an input \emph{set} (often called a \emph{context}) of $K$ elements (called \emph{tokens}) to an output of the same size $K$~\cite{vaswani2017attention}. In most applications, such as language translation, transformers need to map \emph{ordered} sets (i.e. sequences) to other ordered sets, and therefore add special vectors called \emph{positional encodings} to each input element to identify its position in a sequence. These encodings can be learned as part of transformer's training or be hand-crafted.

The most common scheme is the \emph{absolute positional encoding}, where each position in the transformer's $K$-sized context gets a positional vector~\cite{vaswani2017attention}. Some transformers, e.g., \citet{chen2021decision}, use what we call a \emph{global positional encoding}. It is similar to the absolute one, but assigns a separate vector to each position \emph{in the entire input sequence} rather than just the $K$-sized context, up to some maximum length $T \gg K$. Finally, models based on Transformer-XL~\cite{orig,gato,rt2022}, instead condition the attention computation on the \textit{relative} positions between different pairs of input tokens  within a context. 
In this work, we argue that on robotic manipulation finetuning datasets that consist of small numbers of human-gathered demonstrations, relative positional encoding is significantly more data-efficient than absolute or global one.

\vspace{-1mm}
\section{\appr\ architecture and training \label{sec:method}}
\vspace{-1mm}

\vspace{-1mm}
\subsection{Intuition}
\vspace{-1mm}

\begin{wrapfigure}{r!}{0.5\textwidth}
\centering
\includegraphics[width=0.9\linewidth]{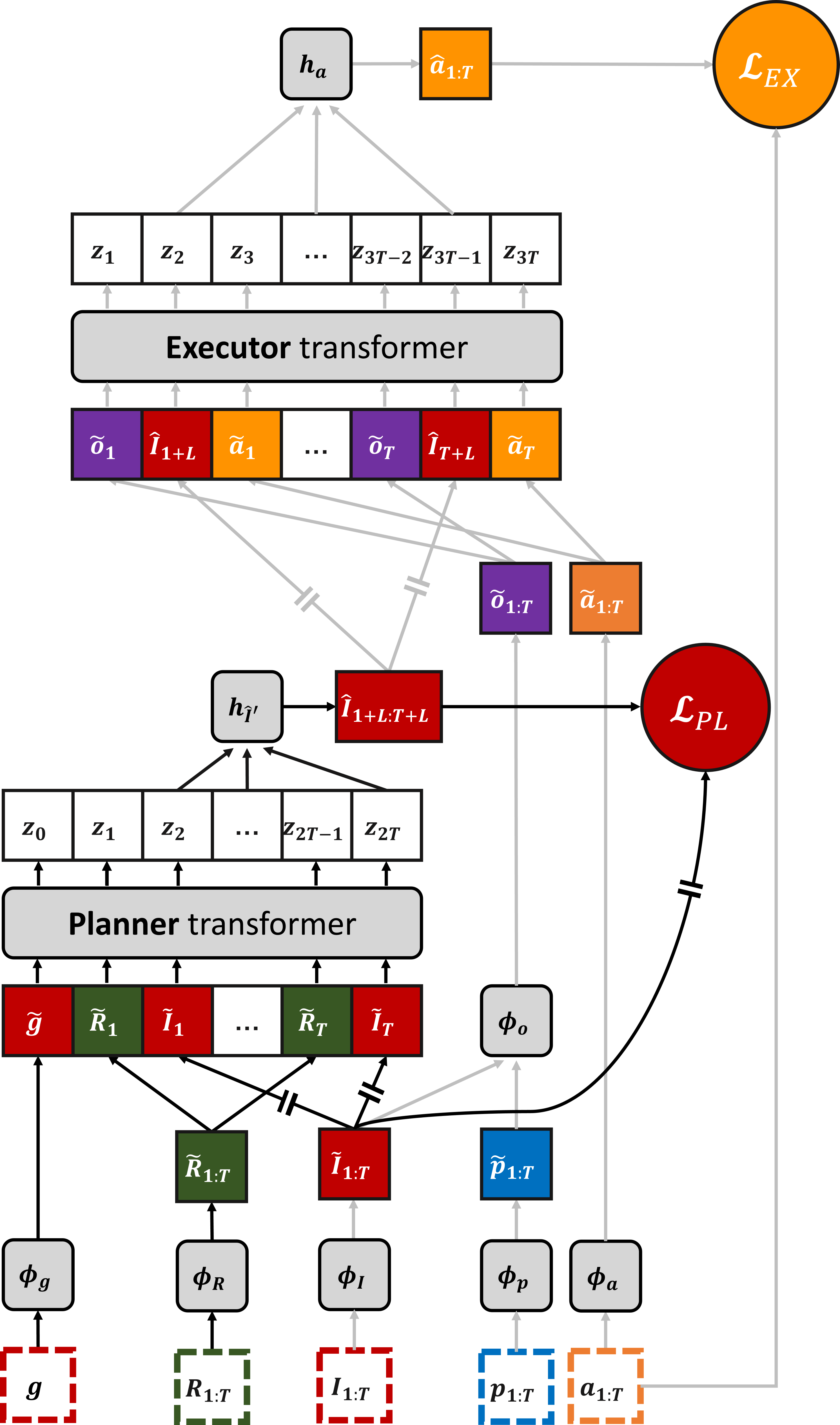}
\caption{\setlength{\parindent}{12pt}
\small \textbf{\appr\ architecture.} \replaced{}{This diagram illustrates the information flow during PLEX training,  described in \Cref{sec:arch}}. \appr\ is optimized using the planner's loss $\LL_{PL}$ (computation shown with black arrows $\uparrow$), and the executor's loss $\LL_{EX}$ (computation shown with gray arrows ${\textcolor{gray} \uparrow}$). The symbols `\textbf{=}' and `\textbf{\textcolor{gray} =}' denote stopgrads, where backpropagation is halted. Each input modality $m$ is embedded using a modality-specific encoder $\phi_m$. Video demonstration embeddings $\tilde g, \tilde I_{1:T}$, and (optionally) $\tilde R_{1:T}$ are used to train the planner \replaced{}{\emph{over the embedding space}} \replaced{with}{using} the prediction loss $\LL_{PL}$ \replaced{in the embedding space}{}. Visuomotor trajectory embeddings $\tilde I_{1:T}, \tilde p_{1:T}, \tilde a_{1:T}$ are passed to the executor to compute the inverse dynamics loss $\LL_{EX}$. Note that if the image encoder $\phi_I$ isn't frozen, $\LL_{EX}$'s gradients will update $\phi_I$. In contrast, the planner's own loss $\LL_{PL}$ never affects $\phi_I$ (see stopgrad symbol \textbf{=}).
}
\label{fig:transformer}
\vspace{-14mm}
\end{wrapfigure}

\appr (shown in \cref{fig:transformer}) separates the model into two transformer-based submodules: 
\begin{enumerate*}[label=\emph{\arabic*)}]
    \item a \emph{planner} that plans in the observation embedding space based on a task specification, and 
    \item an \emph{executor} that takes the embeddings of the historical and the planned future observations and outputs an action to control the robot. 
\end{enumerate*}

This design is motivated by the structure of $\mtvd$, $\vmt$, and $\ttd$ dataset categories, which as we explain below make them suitable for three complementary learning objectives.

\vspace{-1mm}
\begin{enumerate}[leftmargin=*]
    \item  \textbf{Learning to execute state transitions.} The visuomotor trajectories from $\vmt$, collected on the target robotic manipulator or a similar one, show the robot how to {execute} a wide variety of state transitions. By sampling an observation-action tuple $\langle o_{t-H}, \ldots, o_{t}, a_{t}, o_{t+L}\rangle$, the agent can learn to infer $a_{t}$ from $o_{t-H}, \ldots, o_{t}$, and $o_{t+L}$ using \emph{inverse dynamics}, where $t$ is the current time step, $H$ is an observation history length, and $L$ is a lookahead parameter.

    \item \textbf{Learning to plan for tasks.} In order to recommend a meaningful action at each step, inverse dynamics inference needs the (embedding of) the desired future observation. Determining the desired future observation \emph{given a task description} is something that can be learned from multi-task video-only data $\mtvd$, since this data shows what progress towards a successful completion of a specified task should look like.

    \item \textbf{Improving target-task performance.} While learning to plan and execute on diverse $\mtvd$ and $\vmt$ data can result in a robotic manipulation foundation model~\cite{fm2021} with strong zero-shot performance \replaced{}{(see \Cref{sec:gen_exp})}, on many tasks it may be far from perfect. Small datasets $\ttd$ of high-quality target-task demonstrations (e.g., through teleoperation) can provide additional grounding to the target domain to further improve a pretrained model.
\end{enumerate}
\vspace{-1mm}

\vspace{-1mm}
\subsection{Architecture \label{sec:arch}} 
\vspace{-1mm}

Following the above intuitions, we train \appr's executor using data $\vmt$ and \appr's planner using data $\mtvd$, in addition to a small dataset $\ttd$ of target-task trajectories (which, if available, can be used to train both the planner and executor). 
Specifically, let $\tau = g, R_1, I_1, p_1, a_1 \ldots, R_T, I_T, p_T, a_T = g,  R_{1:T}, I_{1:T}, p_{1:T}, a_{1:T}$ denote a trajectory\replaced{, where}{. Here,} $g$ is a task specification, $I_t$ is a tuple of camera image observations, $p_t$ is a proprioceptive state, $a_t$ is an action, and $R_t$ is a return-to-go at time $t$\replaced{; t}{, i.e. $R_t = \sum_{t'=t}^T r_{t'}$, where $r_{t'}$ is the instantaneous reward at time $t'$. T}he length $T$ can vary across trajectories.  As \Cref{fig:transformer} shows, \appr\ processes these input modalities using corresponding encoders $\phi_g$, $\phi_I$, $\phi_p$, $\phi_a$, and $\phi_R$ to obtain an embedded sequence $\tilde g, \tilde R_{1:T}, \tilde I_{1:T}, \tilde p_{1:T}, \tilde a_{1:T}$. 
When a modality is missing, it is replaced by trainable placeholder vectors during embedding. Missing modalities are common in robotic manipulation datasets; e.g., few datasets have rewards. Since \appr's executor and planner are designed to be trainable on task-agnostic visuomotor $\vmt$ data and task-conditioned video-only demonstrations $\mtvd$, respectively, each of these components is specialized to operate only on the (embeddings of) modalities available in their prevalent training data. Per \Cref{fig:transformer}, task description and return embeddings $\tilde g$ and $\tilde R_{1:T}$ don't get routed to the executor, since they are missing from $\vmt$ data. Similarly, the planner only receives $\tilde g$, $\tilde I_{1:T}$ and, optionally, $\tilde R_{1:T}$ embeddings, since they are present in $\mtvd$ data. This separation holds also at deployment time, when all modalities are available.

\vspace{-1mm}
\paragraph{Planner} The planner's sole purpose is to determine \emph{where} the agent should go in the observation embedding space. As shown in \Cref{fig:transformer}, given embeddings $\tilde g, \tilde I_{1:T}$ of a task-conditioned video-only training demonstration, the planner outputs a sequence \replaced{$\hat I_{L:T+L}$}{$\hat I_{1+L:T+L}$} of embeddings corresponding to the observations the agent should ideally see $L$ steps in the future from its current time step; $L$ is a hyperparameter. The planner's training minimizes the prediction loss
\begin{equation}
\textstyle
\LL_{PL}(\tilde g, \tilde R_{1:T}, \tilde I_{1:T}) = \sum_{t=1 + L}^{T + L}\|\tilde I_t- \hat I_t \|_2^2.
\label{eq:pred_loss}
\end{equation}
where we set $\tilde I_t = \tilde I_T$ for $t = T+1, ..., T+L$. 
Crucially, $\LL_{PL}$'s gradients \emph{don't backpropagate} into the encoders $\phi_g$ and $\phi_I$. This is to prevent the collapse of the image embedding space (denoted as $\mathscr{E}_o$); note the stopgrad symbols on $\LL_{PL}$'s computation paths in \Cref{fig:transformer}.
The embedding space $\mathscr{E}_o$ either comes from pretrained encoders or is learned with inverse dynamics during executor training. 

\vspace{-1mm}
\paragraph{Executor}
Like the planner, the executor has a specific role at the deployment time. Given the observation-action sequence $o_{1:t}, a_{1:t}$ so far and the target observation embedding $\hat I_{t+L}$ produced by the planner, the executor infers an action $\hat a_t$ for the current step. This inference step should be done in a task-agnostic way, as the task knowledge  is already incorporated in the $\hat I_{t+L}$ prediction of the planner. For a trajectory from $\vmt$, we optimize the executor via the inverse dynamics loss
\begin{equation}
\textstyle \hspace{-2mm}
\LL_{EX}(I_{1:T}, p_{1:T}, \hat I_{1+L:T+L}, a_{1:T}, ) = \sum_{t=1}^{T-1} \|a_t- \hat a_t \|_2^2
\label{eq:invd_loss}
\end{equation}
A major difference between $\LL_{EX}$ and $\LL_{PL}$ optimization is that the former's gradients can backpropagate into the encoders $\phi_I$, $\phi_o$, $\phi_p$, and $\phi_a$: the computation path for $\LL_{EX}$ through these encoders in \Cref{fig:transformer} doesn't have a stopgrad. This allows executor training to shape the embedding space $\mathscr{E}_o$.

\vspace{-1mm}
\paragraph{Relative positional encoding}
 Like the Decision Transformer (DT)~\cite{chen2021decision}, \appr's planner and executor transformers are derived from GPT-2. However, DT's use of global positional encoding implicitly assumes that all training trajectories have the same length $T$. \appr, in contrast, uses relative encoding from~\citet{orig} as the default. As we show empirically, in robotic manipulation settings where tasks are usually goal-oriented and training demonstrations vary a lot in length, global positional embedding performs poorly and even the fixed absolute positional encoding common in NLP~\cite{vaswani2017attention} performs much better. Especially, for human-collected demonstrations where variability is significant, our experimental results show that relative encoding~\cite{orig} perform significantly better.

\vspace{-1mm}
\subsection{Training \appr}
\vspace{-1mm}

Training \appr\ generally involves both pretraining and finetuning, though the experiments in \Cref{sec:gen_exp} show that pretraining alone already gives \appr\ solid zero-shot performance. 

\textbf{Pretraining} \appr\ consists of two sub-stages:

\emph{1. Pretraining the executor} by optimizing the $\LL_{EX}$ loss (\Cref{eq:invd_loss}) over a $\vmt$ dataset. 

\emph{2. Pretraining the planner} by optimizing the $\LL_{PL}$ loss (\Cref{eq:pred_loss}) over a $\mtvd$ dataset.

If the observation encoders are expected to be trained or finetuned by the inverse dynamics loss $\LL_{EX}$, rather than pretrained and frozen beforehand, it is critical for executor pretraining to be done before training the planner. Indeed, the planner is expected to make predictions in the observation encoders' embedding space, which will change if the inverse dynamics loss affects the encoders. If the encoders are frozen from the start, however, the pretraining stages can proceed asynchronously.

\textbf{Finetuning} involves adapting \appr\ using a target-task demonstration dataset $\ttd$. As with any finetuning, this involves deciding which part of \appr\ to adapt. 

Since $\ttd$  can be viewed both as a small $\mtvd$ and a small $\vmt$ dataset, it can be used to train any component of \appr ---executor, planner, and observation encoders. As with pretraining, if $\ttd$ is used for finetuning the encoders, it is critical to complete their finetuning before finetuning the planner. In \Cref{sec:gen_exp}, we show that finetuning just the last layer of the planner's transformer, which constitutes 5\% of the parameters of the \appr\ instance in the experiment, is sufficient for significantly boosting a pretrained \appr's performance.

$\ttd$ can also be employed for optimizing a behavior cloning loss $\LL_{BC}$. This amounts to \replaced{a stopgrad-free version of $\LL_{EX}$, whose gradients are allowed to backpropagate through the entire \appr\ model, including the planner and/or the encoders}{training the planner, executor, and encoders \emph{simultaneously} by having \appr\ predict  $\ttd$ trajectories's actions from the same trajectories' observations, and allowing the action prediction loss gradients to backpropagate through the entire \appr\ model, to its the inputs}. The experiments in \Cref{sec:enc_exps} demonstrate the efficiency of BC-based finetuning thanks to the use of a relative position encoding.

\vspace{-1mm}
\section{Experiments}
\vspace{-1mm}

We conduct two sets of experiments to answer the following questions: \emph{\textbf{(i)} Does \appr\ pretrained on task-agnostic sensorimotor data and task-annotated video data generalize well to downstream tasks?
\textbf{(ii)} How does the use of relative positional encodings affect \appr's policy quality?} \Cref{app:impl} provides the details about our \appr\ implementation.\footnote{We implement \appr using the GPT-2 of the DT codebase~\cite{chen2021decision} but without return conditioning.}

\vspace{-1mm}
\subsection{Benchmarks and training data}
\vspace{-1mm}

\noindent
\textbf{\mw:} \mw~\cite{mw2019} is a collection of 50 tasks featuring a Sawyer arm. We use \mw-v2 with image observations (see details in \Cref{app:benchmarks}). We consider the ML\replaced{50}{45} split consisting of 45 training and 5 target tasks (\emph{\textbf{door-lock}}, \emph{\textbf{door-unlock}},  \emph{\textbf{hand-insert}},  \emph{\textbf{bin-picking}}, and  \emph{\textbf{box-close}}). We use these 5 target tasks for evaluation. \mw\ comes with high-quality scripted policies for all tasks. To get \emph{\textbf{video demonstration data}} ($\mtvd$), we use these scripted policies to generate 100 successful video-only demonstrations for each of the 45 training tasks, i.e., $|\mtvd|= 4500$.
To generate \emph{\textbf{visuomotor trajectories}} ($\vmt$), for the 5 target tasks' environments, we add zero-mean Gaussian noise with standard deviation $0.5$ to the actions of the scripted policies and record the altered actions. We collect 75 trajectories per task, i.e., $|\vmt| = 375$.
Finally, for \emph{\textbf{target-task demonstrations}} ($\ttd$), we employ the original scripted policies to produce 75 demonstrations per target task and sample 10 of them in a finetuning experiment run, i.e., $|\ttd| = 10$.

\textbf{\rsrw:}  \rsrw\ benchmark~\cite{robosuite}, compared  \mw, has robotic manipulation tasks with a significantly more complicated dynamics and action space. We use 9 of its tasks involving a single robot arm (Panda) (\emph{\textbf{Lift}}, \emph{\textbf{Stack}}, \emph{\textbf{Door}}, \emph{\textbf{NutAssemblyRound}}, \emph{\textbf{NutAssemblySquare}}, \emph{\textbf{PickPlaceBread}} \emph{\textbf{PickPlaceCan}}, \emph{\textbf{PickPlaceMilk}}, and \emph{\textbf{PickPlaceCereal}}). \rsrw's details are provided in \Cref{app:benchmarks}.
Importantly, the training data for \rsrw\ was collected from human demonstrations, \emph{not} generated by scripted policies as in \mw. See \Cref{app:rs_data} for details.

\vspace{-1mm}
\subsection{Generalization experiments \label{sec:gen_exp}}
\vspace{-1mm}

\begin{figure}[t!]
    \hspace{-0.25in}
    \centering
    \includegraphics[width=1\linewidth]{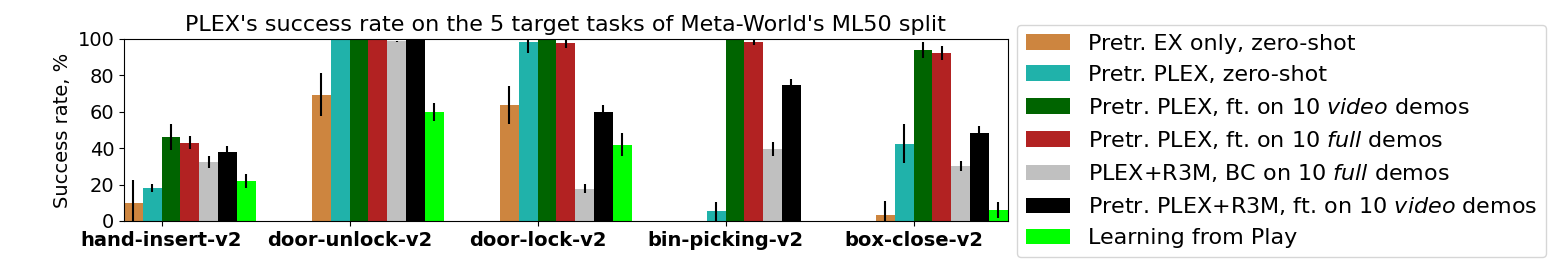}
    \vspace{-0.1in}
    \caption{\appr's generalization experiments. The confidence intervals are computed with 10 seeds.}
    \label{fig:mw}
    \vspace{-0.2in}
\end{figure}

Here we focus on pretraining \appr\ with multi-task \mw\ data. The results are shown in \Cref{fig:mw}. We train a 16,639,149-parameter \appr\ instance (including the ResNet-18-based image encoder) from scratch with random initialization. We use the success rate on the 5 target tasks as the performance metric. For baselines, we experiment with \appr\ with a frozen ResNet-50-based R3M~\cite{nair2022r3m}, an observational representation pretrained on the large Ego4D dataset~\cite{grauman2022ego4d}. We denote it as \emph{\appr{}+R3M}\replaced{.}{; in \Cref{fig:mw}, \emph{Pretr. \appr{}+R3M} was first pretrained on multitask data and then finetuned on a target task, while \emph{\appr{}+R3M, BC} was trained only on a single target task's data from the start.} In addition, we use an adapted \emph{Learning from Play} (LfP) approach~\cite{mcil21rss}. The hyperparameters and details can found \cref{app:impl,app:exp_details}. 
In summary, the experimental results show that \appr\ can perform well without seeing a single sensorimotor expert demonstration.

\vspace{-2mm}
\paragraph{\appr demonstrates \replaced{strong zero-shot performance}{zero-shot generalization capabilities}}
\Cref{fig:mw} shows that \appr\ pretrained on as few as 4500 video demonstrations ($\mtvd$) from the training environments and 250 dynamics trajectories ($\vmt$) from the target environments (denoted as \emph{Pretr. \appr, zero-shot} in \Cref{fig:mw}) exhibits good downstream performance \emph{zero-shot}. 
To demonstrate that this performance is really due to planning learned from video-only data as opposed to the executor inadvertently exploiting biases in the data, we consider a \appr variation (denoted as \emph{Pretr. EX only, zero-shot}) where we only pretrain the \emph{executor} (on $\vmt$), not the planner.\footnote{At run time we feed the embedding of the task's \emph{goal} image as the predictions that the \emph{executor} conditions on (since no planner is trained).} 
The results of \emph{Pretr. EX only, zero-shot} \replaced{shows the baseline}{reflect a level of} performance one can get with knowledge \replaced{}{contained} in the dynamics data $\vmt$ alone. \emph{Pretr. EX only, zero-shot} underperforms  \emph{Pretr. \appr, zero-shot}, which shows the importance of learning from $\mtvd$ via \appr's planner.

\replaced{As a baseline, we also report \emph{Learning from Play} performance, which}{Our main baseline for zero-shot generalization is \emph{Learning from Play} (LfP)~\cite{mcil21rss}, one of the few existing methods able to generalize zero-shot from data as low-quality as $\mtvd$.} LfP has planning capability but doesn't have a way to use either the video-only data $\mtvd$ or the target-task demonstrations $\ttd$. The latter two give \appr\ a large advantage.

\vspace{-2mm}
\paragraph{\replaced{\appr can fast adapt to samll video-only demonstrations}{\appr can be finetuned effectively using only a few video-only demonstrations}}
We further show that finetuning only 5\% of \appr's parameters (the last transformer layer of the planner) on just 10 \emph{video-only} demonstrations for a given task significantly boosts \appr's success rate there. For all 5 downstream tasks, this policy outperforms \emph{Pretr. EX only, zero-shot} by $\geq 2\times$. The improvement is drastic especially in the case of \emph{hand-insert-v2}, \emph{bin-picking-v2}, and \emph{box-close-v2}.

\vspace{-2mm}
\paragraph{Video-only demonstrations is all \appr\ needs during finetuning}
Interestingly, we find that full demonstrations (with both video and action sequences) don't increase \appr's performance beyond video-only ones. This can seen from the experimental results of \emph{Pretr. \appr, ft. on 10 full demos}, where we finetune \appr\ (the action head and last transformer layer of \appr's planner, executor; $\approx11\%$ of \appr) on 10 \emph{full} (sensorimotor) demonstrations for each task. We think this is due to \appr's image encoder being pretrained only on observations from $\vmt$ and frozen during finetuning. Because of this, finetuning couldn't help the encoder learn any extra features for modeling inverse dynamics \emph{over the observation space region covered by $\ttd$}, even if such features would improve \appr's performance.

The issue of impoverished observation coverage in $\vmt$ dataset can be addressed by using a frozen encoder pretrained on an independent large dataset, as the results of \replaced{single-task}{} \emph{\appr{}+R3M, BC} and of \emph{pretrained \appr{}+R3M} in \Cref{fig:mw} suggest. There, \appr's R3M encoder was never trained on \emph{any} \mw\ observations but enables \appr\ to perform reasonably well.

\replaced{ This experiment also shows}{%
The results of  \emph{Pretr. \appr{}+R3M} and \emph{\appr{}+R3M, BC} in \Cref{fig:mw} illuminate} two other aspects of using observation-only representations like R3M: (1) The sensorimotor representation that \appr\ learns \emph{on top of} R3M clearly helps generalization -- \emph{pretrained \appr{}+R3M} performs much better than \replaced{the signle-task one}{\emph{\appr{}+R3M, BC}, which was trained only on a single task's data}, despite \replaced{}{\emph{pretrained \appr{}+R3M}} seeing just video-only demonstrations at finetuning. (2) Fully frozen \replaced{somewhat R3M}{R3M somewhat} limits \appr's performance -- \appr\ \replaced{}{variants} that pretrained \replaced{its}{their} own encoder outperform\replaced{s}{} \appr{}+R3M on 4 of 5 tasks and match it on the remaining one.

\vspace{-1mm}
\subsection{Positional encoding experiments \label{sec:enc_exps}}
\vspace{-1mm}

\begin{figure}[t!]
    \vspace{-0.2in}
    \hspace{-0.25in}
    \includegraphics[width=1.05\linewidth]{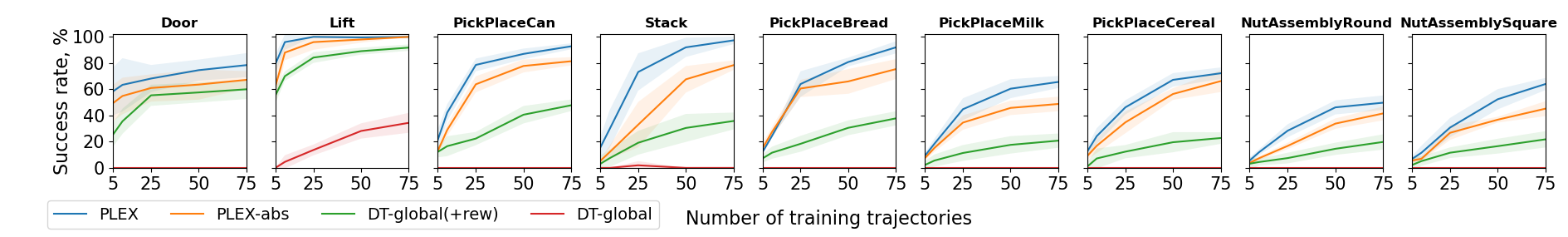}
    \vspace{-0.2in}
    \caption{Data efficiency of \appr's relative positional encoding in single-task mode on \rsrw's single-arm tasks with $|\ttd|$ varying from 5 to 75.  \textbf{\appr} (with relative encodings) in most cases significantly outperforms and at worst matches the performance of its version \textbf{\appr-abs} with absolute positional encodings. Both versions significantly outperform DT.}
    \label{fig:rsrw}
    \vspace{-0.2in}
\end{figure}

In the \mw\ experiments, all training data was generated by scripted policies. In real settings, most such data is generated by people teleoperating robots or performing various tasks themselves. A hallmark of human-generated datasets compared to script-generated ones is the demonstration variability in the former: even trajectories for the same task originating in the same state tend to be different. In this section, we show that in low-data regimes typical of finetuning on human-generated demonstrations, \appr\ with relative positional encoding  yields superior policies for a given amount of training data than using absolute encoding. The results are in \Cref{fig:rsrw}.

\textbf{Baselines, training and evaluation protocol.}  To analyze data efficiency and compare to prior results on \rsrw, we focus on an extreme variant of finetuning -- training from scratch. For each of the 9 \rsrw\ tasks and each of the evaluated encodings, we train a separate 36,242,769-parameter \appr\ instance using only that task's $\ttd$ dataset of full sensorimotor human-generated demonstrations. We compare \appr\ with relative positional encoding to \appr\ with absolute one and to two flavors of the Decision Transformer (DT)~\cite{chen2021decision}, which use global positional embedding.
\Cref{app:rs} and \Cref{fig:rsrw} provide more details about  model training dataset collection, and the baselines. For each task/dataset size/approach, we train on 10 seeds.

\textbf{Results.} As \Cref{fig:rsrw} shows, \appr\ learns strong policies using at most 75 demonstrations, despite having to train a 36M-parameter model including randomly initialized vision models for tasks, most of which have complex dynamics and broad initial state distributions. Moreover, \appr\ with relative positional encoding (denoted simply as \emph{\appr} in the legend) outperforms the alternatives by as much as 20 percentage points (pp) on \rsrw's human-generated demonstration data while never losing to them. In particular, \emph{DT-global(+rew)} and, especially,
\emph{DT-global} perform far worse of both \emph{\appr} and \emph{\appr-abs}. Since all models share most of the implementation and are trained similarly when \emph{\appr} and \emph{\appr-abs} run in BC mode, we attribute \appr's advantage only to the combined effect of using human-generated training data and positional encodings. We have also trained \appr\ and \appr-abs for \mw's 5 target tasks from the previous experiment for various amounts of the available -- scripted -- demonstrations for these tasks and noticed no significant performance difference between \appr\ and \appr-abs on any task. This provides additional evidence that the utility of relative positional enconding manifests itself specifically on human-generated demonstration data.

In fact, relying on relative positional encoding allows \appr\ to achieve state-of-the art performance on all \rsrw\ tasks in this experiment, as we show and analyze empirically in \Cref{app:rs_data}.

\vspace{-1mm}
\section{Conclusion and limitations}
\vspace{-1mm}

We have introduced \appr, a transformer-based sensorimotor model architecture that can be pretrained on robotic manipulation-relevant data realistically available in quantity. Our experimental results show that \appr demonstrate strong zero-shot performance and can be effectively finetuned with demonstrations to further boost its performance. In particular, \appr shows superior performance on human-collected demonstrations because of its usage of relative positional encoding.

\vspace{-2mm}
\paragraph{Limitations}
We believe that \appr\ has great potential as a model architecture for general robotic manipulation, but in most of our experiments so far, the training data came from the same robot on which the trained model was ultimately deployed. In reality, most available multi-task video demonstration data $\mtvd$ is generated by other robots or even people. This can cause a mismatch between the demonstrations and the target robot's capabilities and setups. Planning hierarchically first in the skill space as, e.g., in \citet{lynch2019play}, and then in the observation embedding space may address this issue. In addition, so far we have trained \appr\ on simulated data. The eventual goal, and indeed a significant motivation for this work, would be to pretrain on internet-scale ``in-the-wild'' video datasets~\cite{youtue8m,ho2022imagenvideo,grauman2022ego4d}. Also, with the rise of powerful LLMs such as \citet{instructrl2022}, switching \appr\ to language for task specification can facilitate  generalization across tasks.

\bibliography{corl_paper}

\begin{thebibliography}{58}
\providecommand{\natexlab}[1]{#1}
\providecommand{\url}[1]{\texttt{#1}}
\expandafter\ifx\csname urlstyle\endcsname\relax
  \providecommand{\doi}[1]{doi: #1}\else
  \providecommand{\doi}{doi: \begingroup \urlstyle{rm}\Url}\fi

\bibitem[Vaswani et~al.(2017)Vaswani, Shazeer, Parmar, Uszkoreit, Jones, Gomez,
  Kaiser, and Polosukhin]{vaswani2017attention}
A.~Vaswani, N.~Shazeer, N.~Parmar, J.~Uszkoreit, L.~Jones, A.~N. Gomez,
  L.~Kaiser, and I.~Polosukhin.
\newblock Attention is all you need.
\newblock In \emph{NeurIPS}, 2017.

\bibitem[Brown et~al.(2020)Brown, Mann, Ryder, Subbiah, Kaplan, Dhariwal,
  Neelakantan, Shyam, Sastry, Askell, Agarwal, Herbert-Voss, Krueger, Henighan,
  Child, Ramesh, Ziegler, Wu, Winter, Hesse, Chen, Sigler, Litwin, Gray, Chess,
  Clark, Berner, McCandlish, Radford, Sutskever, and Amodei]{brown2020gpt3}
T.~B. Brown, B.~Mann, N.~Ryder, M.~Subbiah, J.~Kaplan, P.~Dhariwal,
  A.~Neelakantan, P.~Shyam, G.~Sastry, A.~Askell, S.~Agarwal, A.~Herbert-Voss,
  G.~Krueger, T.~Henighan, R.~Child, A.~Ramesh, D.~M. Ziegler, J.~Wu,
  C.~Winter, C.~Hesse, M.~Chen, E.~Sigler, M.~Litwin, S.~Gray, B.~Chess,
  J.~Clark, C.~Berner, S.~McCandlish, A.~Radford, I.~Sutskever, and D.~Amodei.
\newblock Language models are few-shot learners.
\newblock In \emph{NeurIPS}, 2020.

\bibitem[Bommasani et~al.(2021)Bommasani, Hudson, Adeli, Altman, Arora, von
  Arx, Bernstein, Bohg, Bosselut, Brunskill, Brynjolfsson, Buch, Card,
  Castellon, Chatterji, Chen, Creel, Davis, Demszky, Donahue, Doumbouya,
  Durmus, Ermon, Etchemendy, Ethayarajh, Fei-Fei, Finn, Gale, Gillespie, Goel,
  Goodman, Grossman, Guha, Hashimoto, Henderson, Hewitt, Ho, Hong, Hsu, Huang,
  Icard, Jain, Jurafsky, Kalluri, Karamcheti, Keeling, Khani, Khattab, Koh,
  Krass, Krishna, Kuditipudi, Kumar, Ladhak, Lee, Lee, Leskovec, Levent, Li,
  Li, Ma, Malik, Manning, Mirchandani, Mitchell, Munyikwa, Nair, Narayan,
  Narayanan, Newman, Nie, Niebles, Nilforoshan, Nyarko, Ogut, Orr,
  Papadimitriou, Park, Piech, Portelance, Potts, Raghunathan, Reich, Ren, Rong,
  Roohani, Ruiz, Ryan, Ré, Sadigh, Sagawa, Santhanam, Shih, Srinivasan,
  Tamkin, Taori, Thomas, Tramèr, Wang, Wang, Wu, Wu, Wu, Xie, Yasunaga, You,
  Zaharia, Zhang, Zhang, Zhang, Zhang, Zheng, Zhou, and Liang]{fm2021}
R.~Bommasani, D.~A. Hudson, E.~Adeli, R.~Altman, S.~Arora, S.~von Arx, M.~S.
  Bernstein, J.~Bohg, A.~Bosselut, E.~Brunskill, E.~Brynjolfsson, S.~Buch,
  D.~Card, R.~Castellon, N.~Chatterji, A.~Chen, K.~Creel, J.~Q. Davis,
  D.~Demszky, C.~Donahue, M.~Doumbouya, E.~Durmus, S.~Ermon, J.~Etchemendy,
  K.~Ethayarajh, L.~Fei-Fei, C.~Finn, T.~Gale, L.~Gillespie, K.~Goel,
  N.~Goodman, S.~Grossman, N.~Guha, T.~Hashimoto, P.~Henderson, J.~Hewitt,
  D.~E. Ho, J.~Hong, K.~Hsu, J.~Huang, T.~Icard, S.~Jain, D.~Jurafsky,
  P.~Kalluri, S.~Karamcheti, G.~Keeling, F.~Khani, O.~Khattab, P.~W. Koh,
  M.~Krass, R.~Krishna, R.~Kuditipudi, A.~Kumar, F.~Ladhak, M.~Lee, T.~Lee,
  J.~Leskovec, I.~Levent, X.~L. Li, X.~Li, T.~Ma, A.~Malik, C.~D. Manning,
  S.~Mirchandani, E.~Mitchell, Z.~Munyikwa, S.~Nair, A.~Narayan, D.~Narayanan,
  B.~Newman, A.~Nie, J.~C. Niebles, H.~Nilforoshan, J.~Nyarko, G.~Ogut, L.~Orr,
  I.~Papadimitriou, J.~S. Park, C.~Piech, E.~Portelance, C.~Potts,
  A.~Raghunathan, R.~Reich, H.~Ren, F.~Rong, Y.~Roohani, C.~Ruiz, J.~Ryan,
  C.~Ré, D.~Sadigh, S.~Sagawa, K.~Santhanam, A.~Shih, K.~Srinivasan,
  A.~Tamkin, R.~Taori, A.~W. Thomas, F.~Tramèr, R.~E. Wang, W.~Wang, B.~Wu,
  J.~Wu, Y.~Wu, S.~M. Xie, M.~Yasunaga, J.~You, M.~Zaharia, M.~Zhang, T.~Zhang,
  X.~Zhang, Y.~Zhang, L.~Zheng, K.~Zhou, and P.~Liang.
\newblock On the opportunities and risks of foundation models.
\newblock \emph{arXiv}, 2021.

\bibitem[Dasari et~al.(2019)Dasari, Ebert, Tian, Nair, Bucher, Schmeckpeper,
  Singh, Levine, and Finn]{robonetv1}
S.~Dasari, F.~Ebert, S.~Tian, S.~Nair, B.~Bucher, K.~Schmeckpeper, S.~Singh,
  S.~Levine, and C.~Finn.
\newblock Robonet: Large-scale multi-robot learning.
\newblock In \emph{CoRL}, 2019.

\bibitem[Du et~al.(2023)Du, Yang, Dai, Dai, Nachum, Tenenbaum, Schuurmans, and
  Abbeel]{du2023unipi}
Y.~Du, M.~Yang, B.~Dai, H.~Dai, O.~Nachum, J.~B. Tenenbaum, D.~Schuurmans, and
  P.~Abbeel.
\newblock Learning universal policies via text-guided video generation.
\newblock \emph{arXiv}, 2023.

\bibitem[Nair et~al.(2022)Nair, Rajeswaran, Kumar, Finn, and
  Gupta]{nair2022r3m}
S.~Nair, A.~Rajeswaran, V.~Kumar, C.~Finn, and A.~Gupta.
\newblock {R3M}: A universal visual representation for robot manipulation.
\newblock In \emph{CoRL}, 2022.

\bibitem[Dai et~al.(2019)Dai, Yang, Yang, Carbonell, Le, and
  Salakhutdinov]{orig}
Z.~Dai, Z.~Yang, Y.~Yang, J.~G. Carbonell, Q.~V. Le, and R.~Salakhutdinov.
\newblock Transformer-xl: Attentive language models beyond a fixed-length
  context.
\newblock In A.~Korhonen, D.~R. Traum, and L.~M{\`{a}}rquez, editors,
  \emph{ACL}, 2019.

\bibitem[Yen-Chen et~al.(2020)Yen-Chen, Zeng, Song, Isola, and Lin]{yen2020see}
L.~Yen-Chen, A.~Zeng, S.~Song, P.~Isola, and T.-Y. Lin.
\newblock Learning to see before learning to act: Visual pre-training for
  manipulation.
\newblock In \emph{ICRA}, 2020.

\bibitem[Chen et~al.(2021)Chen, Nair, and Finn]{chen2021learning}
A.~S. Chen, S.~Nair, and C.~Finn.
\newblock Learning generalizable robotic reward functions from "in-the-wild"
  human videos.
\newblock In \emph{RSS}, 2021.

\bibitem[Radosavovic et~al.(2022)Radosavovic, Xiao, James, Abbeel, Malik, and
  Darrell]{Radosavovic2022RWRL}
I.~Radosavovic, T.~Xiao, S.~James, P.~Abbeel, J.~Malik, and T.~Darrell.
\newblock Real world robot learning with masked visual pre-training.
\newblock In \emph{CoRL}, 2022.

\bibitem[Lynch and Sermanet(2021)]{mcil21rss}
C.~Lynch and P.~Sermanet.
\newblock Language conditioned imitation learning over unstructured data.
\newblock In \emph{RSS}, 2021.

\bibitem[Jang et~al.(2021)Jang, Irpan, Khansari, Kappler, Ebert, Lynch, Levine,
  and Finn]{jang2021bc}
E.~Jang, A.~Irpan, M.~Khansari, D.~Kappler, F.~Ebert, C.~Lynch, S.~Levine, and
  C.~Finn.
\newblock {BC-Z}: Zero-shot task generalization with robotic imitation
  learning.
\newblock In \emph{Conference on Robot Learning}, pages 991--1002, 2021.

\bibitem[Mandi et~al.(2022)Mandi, Liu, Lee, and Abbeel]{mandi2021generalizable}
Z.~Mandi, F.~Liu, K.~Lee, and P.~Abbeel.
\newblock Towards more generalizable one-shot visual imitation learning.
\newblock In \emph{ICRA}, 2022.

\bibitem[Reed et~al.(2022)Reed, Zolna, Parisotto, Colmenarejo, Novikov,
  Barth-Maron, Gimenez, Sulsky, Kay, Springenberg, Eccles, Bruce, Razavi,
  Edwards, Heess, Chen, Hadsell, Vinyals, Bordbar, and de~Freitas]{gato}
S.~Reed, K.~Zolna, E.~Parisotto, S.~G. Colmenarejo, A.~Novikov, G.~Barth-Maron,
  M.~Gimenez, Y.~Sulsky, J.~Kay, J.~T. Springenberg, T.~Eccles, J.~Bruce,
  A.~Razavi, A.~Edwards, N.~Heess, Y.~Chen, R.~Hadsell, O.~Vinyals, M.~Bordbar,
  and N.~de~Freitas.
\newblock A generalist agent, 2022.

\bibitem[Ebert et~al.(2022)Ebert, Yang, Schmeckpeper, Bucher, Georgakis,
  Daniilidis, Finn, and Levine]{robonetv2}
F.~Ebert, Y.~Yang, K.~Schmeckpeper, B.~Bucher, G.~Georgakis, K.~Daniilidis,
  C.~Finn, and S.~Levine.
\newblock Bridge data: Boosting generalization of robotic skills with
  cross-domain datasets.
\newblock In \emph{RSS}, 2022.

\bibitem[Nasiriany et~al.(2022)Nasiriany, Gao, Mandlekar, and
  Zhu]{nasiriany2022learning}
S.~Nasiriany, T.~Gao, A.~Mandlekar, and Y.~Zhu.
\newblock Learning and retrieval from prior data for skill-based imitation
  learning.
\newblock \emph{arXiv preprint arXiv:2210.11435}, 2022.

\bibitem[Brohan et~al.(2022)Brohan, Brown, Carbajal, Chebotar, Dabis, Finn,
  Gopalakrishnan, Hausman, Herzog, Hsu, Ibarz, Ichter, Irpan, Jackson,
  Jesmonth, Joshi, Julian, Kalashnikov, Kuang, Leal, Lee, Levine, Lu, Malla,
  Manjunath, Mordatch, Nachum, Parada, Peralta, Perez, Pertsch, Quiambao, Rao,
  Ryoo, Salazar, Sanketi, Sayed, Singh, Sontakke, Stone, Tan, Tran, Vanhoucke,
  Vega, Vuong, Xia, Xiao, Xu, Xu, Yu, and Zitkovich]{rt2022}
A.~Brohan, N.~Brown, J.~Carbajal, Y.~Chebotar, J.~Dabis, C.~Finn,
  K.~Gopalakrishnan, K.~Hausman, A.~Herzog, J.~Hsu, J.~Ibarz, B.~Ichter,
  A.~Irpan, T.~Jackson, S.~Jesmonth, N.~J. Joshi, R.~Julian, D.~Kalashnikov,
  Y.~Kuang, I.~Leal, K.-H. Lee, S.~Levine, Y.~Lu, U.~Malla, D.~Manjunath,
  I.~Mordatch, O.~Nachum, C.~Parada, J.~Peralta, E.~Perez, K.~Pertsch,
  J.~Quiambao, K.~Rao, M.~Ryoo, G.~Salazar, P.~Sanketi, K.~Sayed, J.~Singh,
  S.~Sontakke, A.~Stone, C.~Tan, H.~Tran, V.~Vanhoucke, S.~Vega, Q.~Vuong,
  F.~Xia, T.~Xiao, P.~Xu, S.~Xu, T.~Yu, and B.~Zitkovich.
\newblock {RT}-1: Robotics transformer for real-world control at scale.
\newblock \emph{arXiv}, 2022.

\bibitem[Hakhamaneshi et~al.(2021)Hakhamaneshi, Zhao, Zhan, Abbeel, and
  Laskin]{hakhamaneshi2021hierarchical}
K.~Hakhamaneshi, R.~Zhao, A.~Zhan, P.~Abbeel, and M.~Laskin.
\newblock Hierarchical few-shot imitation with skill transition models.
\newblock \emph{arXiv preprint arXiv:2107.08981}, 2021.

\bibitem[Ren et~al.(2021)Ren, Veer, and Majumdar]{ren2021generalization}
A.~Ren, S.~Veer, and A.~Majumdar.
\newblock Generalization guarantees for imitation learning.
\newblock In \emph{Conference on Robot Learning}, pages 1426--1442. PMLR, 2021.

\bibitem[Xihan et~al.(2022)Xihan, Mendez, and Hadfield]{xihan2022skill}
B.~Xihan, O.~Mendez, and S.~Hadfield.
\newblock Skill-il: Disentangling skill and knowledge in multitask imitation
  learning.
\newblock \emph{arXiv preprint arXiv:2205.03130}, 2022.

\bibitem[Mees et~al.(2022)Mees, Hermann, and Burgard]{mees2022matters}
O.~Mees, L.~Hermann, and W.~Burgard.
\newblock What matters in language conditioned robotic imitation learning over
  unstructured data.
\newblock \emph{IEEE Robotics and Automation Letters}, 7\penalty0 (4):\penalty0
  11205--11212, 2022.

\bibitem[Nair et~al.(2017)Nair, Chen, Agrawal, Isola, Abbeel, Malik, and
  Levine]{nair2017knot}
A.~Nair, D.~Chen, P.~Agrawal, P.~Isola, P.~Abbeel, J.~Malik, and S.~Levine.
\newblock Combining self-supervised learning and imitation for vision-based
  rope manipulation.
\newblock In \emph{ICRA}, 2017.

\bibitem[Radosavovic et~al.(2021)Radosavovic, Wang, Pinto, and
  Malik]{rados2021soil}
I.~Radosavovic, X.~Wang, L.~Pinto, and J.~Malik.
\newblock State-only imitation learning for dexterous manipulation.
\newblock In \emph{IROS}, 2021.

\bibitem[Pathak et~al.(2018)Pathak, Mahmoudieh, Luo, Agrawal, Chen, Shentu,
  Shelhamer, Malik, Efros, and Darrell]{pathak2018zs}
D.~Pathak, P.~Mahmoudieh, G.~Luo, P.~Agrawal, D.~Chen, Y.~Shentu, E.~Shelhamer,
  J.~Malik, A.~A. Efros, and T.~Darrell.
\newblock Zero-shot visual imitation.
\newblock In \emph{ICLR}, 2018.

\bibitem[Baker et~al.(2022)Baker, Akkaya, Zhokhov, Huizinga, Tang, Ecoffet,
  Houghton, Sampedro, and Clune]{vpt2022}
B.~Baker, I.~Akkaya, P.~Zhokhov, J.~Huizinga, J.~Tang, A.~Ecoffet, B.~Houghton,
  R.~Sampedro, and J.~Clune.
\newblock Video pretraining (vpt): Learning to act by watching unlabeled online
  videos.
\newblock In \emph{NeurIPS}, 2022.

\bibitem[Xu et~al.(2022)Xu, Jiang, Li, and Zhan]{xu2022pgiorl}
H.~Xu, L.~Jiang, J.~Li, and X.~Zhan.
\newblock A policy-guided imitation approach for offline reinforcement
  learning.
\newblock In \emph{arXiv}, 2022.

\bibitem[Dasari and Gupta(2020)]{dasari2020transformers}
S.~Dasari and A.~Gupta.
\newblock Transformers for one-shot visual imitation.
\newblock In \emph{CoRL}, 2020.

\bibitem[Kim et~al.(2021)Kim, Ohmura, and Kuniyoshi]{kim2021transformer}
H.~Kim, Y.~Ohmura, and Y.~Kuniyoshi.
\newblock Transformer-based deep imitation learning for dual-arm robot
  manipulation.
\newblock In \emph{2021 IEEE/RSJ International Conference on Intelligent Robots
  and Systems (IROS)}, pages 8965--8972. IEEE, 2021.

\bibitem[Ho et~al.(2022)Ho, Chan, Saharia, Whang, Gao, Gritsenko, Kingma,
  Poole, Norouzi, Fleet, and Salimans]{ho2022imagenvideo}
J.~Ho, W.~Chan, C.~Saharia, J.~Whang, R.~Gao, A.~Gritsenko, D.~P. Kingma,
  B.~Poole, M.~Norouzi, D.~J. Fleet, and T.~Salimans.
\newblock Imagen {V}ideo: High definition video generation with diffusion
  models.
\newblock \emph{arXiv}, 2022.

\bibitem[Yu et~al.(2019)Yu, Quillen, He, Julian, Narayan, Shively, Bellathur,
  Hausman, Finn, and Levine]{mw2019}
T.~Yu, D.~Quillen, Z.~He, R.~Julian, A.~Narayan, H.~Shively, A.~Bellathur,
  K.~Hausman, C.~Finn, and S.~Levine.
\newblock Meta-world: A benchmark and evaluation for multi-task and meta
  reinforcement learning.
\newblock In \emph{CoRL}, 2019.

\bibitem[Zhu et~al.(2020)Zhu, Wong, Mandlekar, and Martín-Martín]{robosuite}
Y.~Zhu, J.~Wong, A.~Mandlekar, and R.~Martín-Martín.
\newblock Robosuite: A modular simulation framework and benchmark for robot
  learning, 2020.

\bibitem[Mandlekar et~al.(2021)Mandlekar, Xu, Wong, Nasiriany, Wang, Kulkarni,
  Fei-Fei, Savarese, Zhu, and Mart\'{i}n-Mart\'{i}n]{robomimic2021}
A.~Mandlekar, D.~Xu, J.~Wong, S.~Nasiriany, C.~Wang, R.~Kulkarni, L.~Fei-Fei,
  S.~Savarese, Y.~Zhu, and R.~Mart\'{i}n-Mart\'{i}n.
\newblock What matters in learning from offline human demonstrations for robot
  manipulation.
\newblock In \emph{CoRL}, 2021.

\bibitem[Grauman et~al.(2022)Grauman, Westbury, Byrne, Chavis, Furnari,
  Girdhar, Hamburger, Jiang, Liu, Liu, Martin, Nagarajan, Radosavovic,
  Ramakrishnan, Ryan, Sharma, Wray, Xu, Xu, Zhao, Bansal, Batra, Cartillier,
  Crane, Do, Doulaty, Erapalli, Feichtenhofer, Fragomeni, Fu, Gebreselasie,
  Gonzalez, Hillis, Huang, Huang, Jia, Khoo, Kolar, Kottur, Kumar, Landini, Li,
  Li, Li, Mangalam, Modhugu, Munro, Murrell, Nishiyasu, Price, Puentes,
  Ramazanova, Sari, Somasundaram, Southerland, Sugano, Tao, Vo, Wang, Wu, Yagi,
  Zhao, Zhu, Arbelaez, Crandall, Damen, Farinella, Fuegen, Ghanem, Ithapu,
  Jawahar, Joo, Kitani, Li, Newcombe, Oliva, Park, Rehg, Sato, Shi, Shou,
  Torralba, Torresani, Yan, and Malik]{grauman2022ego4d}
K.~Grauman, A.~Westbury, E.~Byrne, Z.~Chavis, A.~Furnari, R.~Girdhar,
  J.~Hamburger, H.~Jiang, M.~Liu, X.~Liu, M.~Martin, T.~Nagarajan,
  I.~Radosavovic, S.~K. Ramakrishnan, F.~Ryan, J.~Sharma, M.~Wray, M.~Xu, E.~Z.
  Xu, C.~Zhao, S.~Bansal, D.~Batra, V.~Cartillier, S.~Crane, T.~Do, M.~Doulaty,
  A.~Erapalli, C.~Feichtenhofer, A.~Fragomeni, Q.~Fu, A.~Gebreselasie,
  C.~Gonzalez, J.~Hillis, X.~Huang, Y.~Huang, W.~Jia, W.~Khoo, J.~Kolar,
  S.~Kottur, A.~Kumar, F.~Landini, C.~Li, Y.~Li, Z.~Li, K.~Mangalam,
  R.~Modhugu, J.~Munro, T.~Murrell, T.~Nishiyasu, W.~Price, P.~R. Puentes,
  M.~Ramazanova, L.~Sari, K.~Somasundaram, A.~Southerland, Y.~Sugano, R.~Tao,
  M.~Vo, Y.~Wang, X.~Wu, T.~Yagi, Z.~Zhao, Y.~Zhu, P.~Arbelaez, D.~Crandall,
  D.~Damen, G.~M. Farinella, C.~Fuegen, B.~Ghanem, V.~K. Ithapu, C.~V. Jawahar,
  H.~Joo, K.~Kitani, H.~Li, R.~Newcombe, A.~Oliva, H.~S. Park, J.~M. Rehg,
  Y.~Sato, J.~Shi, M.~Z. Shou, A.~Torralba, L.~Torresani, M.~Yan, and J.~Malik.
\newblock Ego4d: Around the world in 3,000 hours of egocentric video, 2022.

\bibitem[Damen et~al.(2018)Damen, Doughty, Farinella, Fidler, Furnari, Kazakos,
  Moltisanti, Munro, Perrett, Price, and Wray]{Damen2018EPICKITCHENS}
D.~Damen, H.~Doughty, G.~M. Farinella, S.~Fidler, A.~Furnari, E.~Kazakos,
  D.~Moltisanti, J.~Munro, T.~Perrett, W.~Price, and M.~Wray.
\newblock Scaling egocentric vision: The epic-kitchens dataset.
\newblock In \emph{European Conference on Computer Vision (ECCV)}, 2018.

\bibitem[Damen et~al.(2022)Damen, Doughty, Farinella, , Furnari, Ma, Kazakos,
  Moltisanti, Munro, Perrett, Price, and Wray]{Damen2022RESCALING}
D.~Damen, H.~Doughty, G.~M. Farinella, , A.~Furnari, J.~Ma, E.~Kazakos,
  D.~Moltisanti, J.~Munro, T.~Perrett, W.~Price, and M.~Wray.
\newblock Rescaling egocentric vision: Collection, pipeline and challenges for
  epic-kitchens-100.
\newblock \emph{International Journal of Computer Vision (IJCV)}, 130:\penalty0
  33–55, 2022.

\bibitem[Abu-El-Haija et~al.(2016)Abu-El-Haija, Kothari, Lee, Natsev, Toderici,
  Varadarajan, and Vijayanarasimhan]{youtue8m}
S.~Abu-El-Haija, N.~Kothari, J.~Lee, P.~Natsev, G.~Toderici, B.~Varadarajan,
  and S.~Vijayanarasimhan.
\newblock Youtube-8m: A large-scale video classification benchmark, 2016.

\bibitem[Chen et~al.(2021)Chen, Lu, Rajeswaran, Lee, Grover, Laskin, Abbeel,
  Srinivas, and Mordatch]{chen2021decision}
L.~Chen, K.~Lu, A.~Rajeswaran, K.~Lee, A.~Grover, M.~Laskin, P.~Abbeel,
  A.~Srinivas, and I.~Mordatch.
\newblock Decision transformer: Reinforcement learning via sequence modeling.
\newblock In \emph{NeurIPS}, 2021.

\bibitem[Lynch et~al.(2019)Lynch, Khansari, Xiao, Kumar, Tompson, Levine, and
  Sermanet]{lynch2019play}
C.~Lynch, M.~Khansari, T.~Xiao, V.~Kumar, J.~Tompson, S.~Levine, and
  P.~Sermanet.
\newblock Learning latent plans from play.
\newblock In \emph{coRL}, 2019.

\bibitem[Ouyang et~al.(2022)Ouyang, Wu, Jiang, Almeida, Wainwright, Mishkin,
  Zhang, Agarwal, Slama, Ray, Schulman, Hilton, Kelton, Miller, Simens, Askell,
  Welinder, Christiano, Leike, and Lowe]{instructrl2022}
L.~Ouyang, J.~Wu, X.~Jiang, D.~Almeida, C.~L. Wainwright, P.~Mishkin, C.~Zhang,
  S.~Agarwal, K.~Slama, A.~Ray, J.~Schulman, J.~Hilton, F.~Kelton, L.~Miller,
  M.~Simens, A.~Askell, P.~Welinder, P.~Christiano, J.~Leike, and R.~Lowe.
\newblock Training language models to follow instructions with human feedback.
\newblock \emph{arXiv preprint arXiv:2203.02155}, 2022.

\bibitem[Chane-Sane et~al.(2023)Chane-Sane, Schmid, and
  Laptev]{chane2023learning}
E.~Chane-Sane, C.~Schmid, and I.~Laptev.
\newblock Learning video-conditioned policies for unseen manipulation tasks.
\newblock \emph{arXiv preprint arXiv:2305.06289}, 2023.

\bibitem[Karamcheti et~al.(2023)Karamcheti, Nair, Chen, Kollar, Finn, Sadigh,
  and Liang]{karamcheti2023language}
S.~Karamcheti, S.~Nair, A.~S. Chen, T.~Kollar, C.~Finn, D.~Sadigh, and
  P.~Liang.
\newblock Language-driven representation learning for robotics.
\newblock \emph{arXiv preprint arXiv:2302.12766}, 2023.

\bibitem[Yu et~al.(2018)Yu, Finn, Xie, Dasari, Zhang, Abbeel, and
  Levine]{yu2018daml}
T.~Yu, C.~Finn, A.~Xie, S.~Dasari, T.~Zhang, P.~Abbeel, and S.~Levine.
\newblock One-shot imitation from observing humans via domain-adaptive
  meta-learning.
\newblock In \emph{RSS}, 2018.

\bibitem[Zhou et~al.(2020)Zhou, Jang, Kappler, Herzog, Khansari, Wohlhart, Bai,
  Kalakrishnan, Levine, and Finn]{zhou2020watch}
A.~Zhou, E.~Jang, D.~Kappler, A.~Herzog, M.~Khansari, P.~Wohlhart, Y.~Bai,
  M.~Kalakrishnan, S.~Levine, and C.~Finn.
\newblock Watch, try, learn: Meta-learning from demonstrations and reward.
\newblock In \emph{ICLR}, 2020.

\bibitem[Li et~al.(2022)Li, Lu, Cao, Cai, and Wang]{li2022meta}
J.~Li, T.~Lu, X.~Cao, Y.~Cai, and S.~Wang.
\newblock Meta-imitation learning by watching video demonstrations.
\newblock In \emph{International Conference on Learning Representations}, 2022.

\bibitem[Finn et~al.(2017)Finn, Yu, Zhang, Abbeel, and Levine]{finn2017os}
C.~Finn, T.~Yu, T.~Zhang, P.~Abbeel, and S.~Levine.
\newblock One-shot visual imitation learning via meta-learning.
\newblock In \emph{CoRL}, 2017.

\bibitem[Mandi et~al.(2022)Mandi, Abbeel, and James]{mandi2022effectiveness}
Z.~Mandi, P.~Abbeel, and S.~James.
\newblock On the effectiveness of fine-tuning versus meta-reinforcement
  learning.
\newblock \emph{CoRL 2022 Workshop on Pre-training Robot Learning}, 2022.

\bibitem[Singh et~al.(2020)Singh, Yu, Yang, Zhang, Kumar, and
  Levine]{singh2020cog}
A.~Singh, A.~Yu, J.~Yang, J.~Zhang, A.~Kumar, and S.~Levine.
\newblock Cog: Connecting new skills to past experience with offline
  reinforcement learning.
\newblock In \emph{CoRL}, 2020.

\bibitem[Schmeckpeper et~al.(2020)Schmeckpeper, Rybkin, Daniilidis, Levine, and
  Finn]{schmeckpeper2020reinforcement}
K.~Schmeckpeper, O.~Rybkin, K.~Daniilidis, S.~Levine, and C.~Finn.
\newblock Reinforcement learning with videos: Combining offline observations
  with interaction.
\newblock In \emph{CoRL}, 2020.

\bibitem[Venuto et~al.(2022)Venuto, Yang, Abbeel, Precup, Mordatch, and
  Nachum]{venuto2022multi}
D.~Venuto, S.~Yang, P.~Abbeel, D.~Precup, I.~Mordatch, and O.~Nachum.
\newblock Multi-environment pretraining enables transfer to action limited
  datasets.
\newblock \emph{arXiv preprint arXiv:2211.13337}, 2022.

\bibitem[Dosovitskiy et~al.(2021)Dosovitskiy, Beyer, Kolesnikov, Weissenborn,
  Zhai, Unterthiner, Dehghani, Minderer, Heigold, Gelly, Uszkoreit, and
  Houlsby]{dosovitskiy2021image}
A.~Dosovitskiy, L.~Beyer, A.~Kolesnikov, D.~Weissenborn, X.~Zhai,
  T.~Unterthiner, M.~Dehghani, M.~Minderer, G.~Heigold, S.~Gelly, J.~Uszkoreit,
  and N.~Houlsby.
\newblock An image is worth 16x16 words: Transformers for image recognition at
  scale.
\newblock In \emph{ICLR}, 2021.

\bibitem[Janner et~al.(2021)Janner, Li, and Levine]{janner2021reinforcement}
M.~Janner, Q.~Li, and S.~Levine.
\newblock Reinforcement learning as one big sequence modeling problem, 2021.

\bibitem[Shang et~al.(2021)Shang, Kahatapitiya, Li, and
  Ryoo]{shang2021starformer}
J.~Shang, K.~Kahatapitiya, X.~Li, and M.~S. Ryoo.
\newblock Starformer: Transformer with state-action-reward representations for
  visual reinforcement learning.
\newblock \emph{arXiv preprint arXiv:2110.06206}, 2021.

\bibitem[Loynd et~al.(2020)Loynd, Fernandez, Celikyilmaz, Swaminathan, and
  Hausknecht]{WMG}
R.~Loynd, R.~Fernandez, A.~Celikyilmaz, A.~Swaminathan, and M.~Hausknecht.
\newblock Working memory graphs.
\newblock In \emph{ICML}, 2020.

\bibitem[Parisotto et~al.(2020)Parisotto, Song, Rae, Pascanu, Gulcehre,
  Jayakumar, Jaderberg, Kaufman, Clark, Noury, Botvinick, Heess, and
  Hadsell]{GTrXL}
E.~Parisotto, F.~Song, J.~Rae, R.~Pascanu, C.~Gulcehre, S.~Jayakumar,
  M.~Jaderberg, R.~L. Kaufman, A.~Clark, S.~Noury, M.~Botvinick, N.~Heess, and
  R.~Hadsell.
\newblock Stabilizing transformers for reinforcement learning.
\newblock In \emph{ICML}, 2020.

\bibitem[Dance et~al.(2021)Dance, Perez, and Cachet]{dance2021conditioned}
C.~R. Dance, J.~Perez, and T.~Cachet.
\newblock Conditioned reinforcement learning for few-shot imitation.
\newblock In \emph{International Conference on Machine Learning}, pages
  2376--2387. PMLR, 2021.

\bibitem[Ahn et~al.(2022)Ahn, Brohan, Brown, Chebotar, Cortes, David, Finn,
  Gopalakrishnan, Hausman, Herzog, et~al.]{ahn2022can}
M.~Ahn, A.~Brohan, N.~Brown, Y.~Chebotar, O.~Cortes, B.~David, C.~Finn,
  K.~Gopalakrishnan, K.~Hausman, A.~Herzog, et~al.
\newblock Do as {I} can, not as {I} say: Grounding language in robotic
  affordances.
\newblock In \emph{CoRL}, 2022.

\bibitem[He et~al.(2016)He, Zhang, Ren, and Sun]{resnet}
K.~He, X.~Zhang, S.~Ren, and J.~Sun.
\newblock Deep residual learning for image recognition.
\newblock In \emph{CVPR}, 2016.

\bibitem[Levine et~al.(2016)Levine, Finn, Darrell, and Abbeel]{levine2016end}
S.~Levine, C.~Finn, T.~Darrell, and P.~Abbeel.
\newblock End-to-end training of deep visuomotor policies.
\newblock \emph{The Journal of Machine Learning Research}, 17\penalty0
  (1):\penalty0 1334--1373, 2016.

\end{thebibliography}

\newpage
\onecolumn
\begin{appendices}

\section{Related work \label[appendix]{app:rel_work}}

Our work lies at the intersection of scalable multi-task representation learning for robotic manipulation, learning from observations, and decision-making using transformers.

\textbf{Representation learning for robotic manipulation.} Most approaches of this kind focus on pretraining purely \emph{non-motor}, usually visual, representation models (see, e.g., \cite{yen2020see,chen2021learning,nair2022r3m,chane2023learning,karamcheti2023language}, and references therein). These models don't output actions; they are meant to be foundations on top of which a policy network is to be learned. Thus, in contrast to \appr, by themselves they can't enable zero-shot generalization to unseen tasks even in the limit of pretraining data coverage and amount. However, they are synergistic with \appr: \appr\ can use them as frozen observation encoders, as we show in \Cref{sec:gen_exp} on the example of R3M~\cite{nair2022r3m}.

Techniques that train sensorimotor models -- i.e., full-fledged generalist policies, like \appr\ -- have also been rising in prominence. Some of them~\cite{yu2018daml,zhou2020watch,li2022meta} are based on \emph{meta learning}~\cite{finn2017os}. However, \citet{mandi2022effectiveness} have shown multi-task pretraining followed by finetuning to be more effective when the task distribution is broad, and several approaches~\cite{mcil21rss,jang2021bc,mandi2021generalizable,gato,robonetv2,nasiriany2022learning,rt2022} follow this training paradigm as does \appr. At the same time, most of them need pretraining data consisting of high-quality demonstrations in the form of matching videos \emph{and} action sequences. While the quality requirement can be relaxed using offline RL, as, e.g., in \citet{singh2020cog}, in order to enable generalization across broad task distributions these sensorimotor training demonstrations need correspondingly broad task coverage. This assumption is presently unrealistic and ignores the vast potential of the available video-only data --- the weakness \appr\ aims to address. 

Among the sensorimotor representation learning methods that, like \appr, try to learn from both video-only and sensorimotor data are \citet{schmeckpeper2020reinforcement}, \citet{mcil21rss}, and \citet{mees2022matters}. \citet{schmeckpeper2020reinforcement} consider single-task settings only and require the video-only and sensorimotor data to provide demonstrations for the same tasks. \citet{mcil21rss} and \citet{mees2022matters} allow the sensorimotor data to come from exploratory policies rather than task demonstrations but insist that this data must be generated from meaningful \emph{skills}, a strong assumption that \appr\ avoids. 

Architecturally, most aforementioned approaches use monolithic models that don't have separate components for planning and execution like \appr. Notable exceptions are methods that mine skills from pretraining data, embed them into a latent space, and use the latent skill space for accelerated policy learning of new tasks after pretraining~\cite{hakhamaneshi2021hierarchical,ren2021generalization,xihan2022skill,nasiriany2022learning,mees2022matters}. This is akin to planning in the skill space. 
\appr\ can accommodate this approach hierarchically by having, e.g., a CVAE-based high-level planning model~\cite{lynch2019play} produce a task-conditioned sequence of skill latents and feeding them into a skill-conditioned planning model that will plan in the observation embedding space. However, in this work's experiments, for simplicity \appr\ plans in the observation embedding space directly.

\textbf{Learning and imitation from observations (I/LfO)} I/LfO has been used in robotic manipulation both for single-task tabula-rasa policy learning~\cite{nair2017knot,rados2021soil} and pretraining~\cite{pathak2018zs}. \citet{pathak2018zs} is related to \appr\ in spirit but lacks a counterpart of \appr's planner. As a result, it can't complete an unseen task based on the task's goal description alone: it needs either a sequence of subgoal images starting at the robot's initial state or a sequence of landmarks common to all initial states of a given task. Beyond robotics, a type of LfO was also employed by~\citet{vpt2022} and \citet{venuto2022multi} to pretrain a large sensorimotor model for Minecraft and Atari, respectively. This model, like \citet{pathak2018zs}'s, doesn't have a task-conditioned planning capability and is meant to serve only as a finetunable behavioral prior. \citet{xu2022pgiorl} investigate an LfO method akin to \appr\ in low-dimensional environments, where it side-steps the question of choosing an appropriate representation for planning, the associated efficiency tradeoffs, and pretraining a generalizable planning policy.

Overall, the closest approach to \appr\ is the concurrently proposed UniPi~\cite{du2023unipi}. It also has a universal planner meant to be pretrained on a large collection of available videos, as well as an executor that captures inverse dynamics. However, UniPi ignores the issue of data efficiency and plans in the space of images (observations), using diffusion~\cite{ho2022imagenvideo}, rather than in the latent space of their embeddings. This is expensive to learn and potentially detrimental to plan quality.  Latents even from statically pretrained image encoders are sufficient to capture object manipulation-relevant details from videoframes~\cite{yen2020see}, whereas diffusion models can easily miss these details or model their 3D structure inconsistently~\cite{ho2022imagenvideo}. Indeed, despite being conceptually capable of closed-loop control, for computational efficiency reasons UniPi generates open-loop plans, while \appr\ interleaves planning and execution in a closed loop. 

\textbf{Transformers for decision making and their data efficiency.} After emerging as the dominant paradigm in NLP~\cite{brown2020gpt3} and CV~\cite{dosovitskiy2021image}, transformers have
been recently applied to solving general long-horizon decision-making problems by imitation and reinforcement learning~\cite{chen2021decision,janner2021reinforcement,shang2021starformer,WMG,GTrXL}, including multi-task settings~\cite{dance2021conditioned} and robotic manipulation~\cite{dasari2020transformers,kim2021transformer,mees2022matters,gato,rt2022}. \citet{mees2022matters} provide evidence that in robotic manipulation transformers perform better than RNNs~\cite{mcil21rss} while having many fewer parameters. Of all these works, only \citet{gato} uses relative positional encoding, and only by ``inheriting'' it with the overall Transformer-XL architecture~\cite{orig}, without motivating its effectiveness for decision-making.

\textbf{Task specification formats.} Task specification modality can significantly influence the generalization power of models pretrained on multi-task data. Common task conditioning choices are images of a task's goal~\cite{robonetv2}, videos of a task demonstration by a person~\cite{yu2018daml,jang2021bc} or by a robot~\cite{finn2017os,mandi2021generalizable}, and language descriptions~\cite{mcil21rss,jang2021bc,mees2022matters,ahn2022can,rt2022}. \appr\ is compatible with any of these formats; in the experiments, we use goal images.

\section{Problem formalization \label[appendix]{app:pomdp}}

Formally, the problem \appr\ aims to solve can be described as a partially observable Markov decision process (POMDP) $\langle \GG, \SS, \OO, z, \AA, p, r\rangle$ with a special structure. Here, $\GG$ is the space of possible manipulation tasks that we may want to carry out the tasks in $\GG$. $\SS = \PP \times \WW$ is a state space consisting of a space $\PP$ of robots' proprioceptive states (e.g., poses, joint speeds, etc.) and a space $\WW$ of world states. A state $s$'s proprioceptive part $p \in \PP$ is known at execution time and in some of the training data, whereas the world state $w \in \WW$ is never observable directly. A latent state $s$ can be probabilistically inferred from its observations $o\in \OO$ and a state-conditioned distribution $z: \SS \rightarrow \Delta(\OO)$ that describes how latent states in $\SS$ manifest themselves through observations, where $\Delta$ denotes the space of distributions. For robotic manipulation, each observation can consist of several \emph{modalities}: camera images (possibly from several cameras at each time step), depth maps, tactile sensor readings, etc. The distribution $z$ is unknown and needs to be learned. $\AA$ is an action space, e.g., the space of all pose changes the robotic manipulator can achieve in 1 time step, and $p: \SS \times \AA \rightarrow \Delta (\SS)$ is a transition function describing how executing an action affects a current state, which potentially is stochastic. A reward function $r: \GG \times \SS \times \AA \times \SS  \rightarrow \mathbb{R}$ can provide additional detail about task execution by assigning a numeric reward to each state transition, e.g., 0 for transitions to a task's goal state and -1 otherwise. Our objective is to learn a policy $\pi: \GG \times \OO_{|H} \rightarrow A$ that maps a history of observations $\OO_{|H}$ over the previous $H$ steps to an action so as to lead the robot to accomplish a task $g \in \GG$.

\section{\appr\ implementation details \label[appendix]{app:impl}}

The transformers \appr\ uses as its planner and executor are derived from the GPT-2-based version of the Decision Transformer (DT)~\cite{chen2021decision}. Like in DT, we feed inputs into \appr\ by embedding each modality instance (e.g., an image or an action) as a single unit. This is different to the way, e.g., Gato~\cite{gato} and Trajectory Transformer~\cite{janner2021reinforcement} do it, by splitting each input into fragments such as image patches and embedding each fragment separately.

 We condition \appr's planner on embeddings of goal images. Low-dimensional inputs (actions and proprioceptive states) are mapped to $\mathbb{R}^h$, the transformer's $h$-dimensional input space, using a 1-layer linear neural network. High-dimensional inputs -- videoframes from one or several cameras at each time step as well as goal images -- are processed using a ResNet-18-based~\cite{resnet} encoder from Robomimic~\cite{robomimic2021}. It applies a random crop augmentation to each camera's image, passes it through a separate ResNet18 instance associated with that camera, then passes the result through a spatial softmax layer \cite{levine2016end}, and finally through a small MLP. The resulting embedding is fed into \appr's planner. If the robot has several cameras, the encoder has a separate ResNet instance for each. For each time step, \appr's planner outputs an $h$-dimensional latent state representing the \emph{predicted} embedding of \appr's visual observations $k$ time steps into the future, where $k$ is a tunable parameter. These latents are then fed directly into the planner as predictions of future observation embeddings. The output latents from the planner transformer are fed through a $\tanh$ non-linearity, which outputs action vectors in the $[-1, 1]$ range. The hyperparameters can be found in \Cref{tab:arch_params,tab:training_params}.

Our \appr\ implementation is available at \href{https://microsoft.github.io/PLEX}{https://microsoft.github.io/PLEX}.

\section{Additional details about the experiments \label[appendix]{app:exp_details}}

\subsection{\mw\ and \rsrw\ details \label[appendix]{app:benchmarks}}
\textbf{\mw.} In our \mw-v2 setup, at each time step the agent receives an $84 \times 84$ image from the environment's \emph{corner} camera and the Sawyer arm's 18D proprioceptive state. The agent's actions have 4 dimensions, each scaled to the $[-1, 1]$ range. Although \mw\ also provides privileged information about the state of the environment, including the poses of all relevant objects, our \appr\ agent doesn't access it.

\textbf{\rsrw.} The observation and action space in our experiments is exactly as in the best-performing high-dimensional setup from the Robomimic paper~\cite{robomimic2021}. Namely, actions are 7-dimensional: 6 dimensions for the gripper's pose control (OSC\_POSE) and 1 for opening/closing it. Visual observations are a pair of $84 \times 84$ images from \emph{agentview} (frontal) and \emph{eye-in-hand} (wrist) cameras at each step. Proprioceptive states consist of a 3D gripper position, a 4D quaternion for its orientation, and 2D gripper fingers' position.

\subsection{Details of the baselines from prior work \label[appendix]{app:baseline_details}}

\textbf{\appr+R3M~\cite{nair2022r3m}.} We experiment with two combinations of \appr\ with a frozen ResNet-50-based R3M~\cite{nair2022r3m}, an observational representation pretrained on the large Ego4D dataset~\cite{grauman2022ego4d} In these experiments, R3M replaces Robomimic's ResNet-18, and we use versions of our \mw\ $\vmt$, $\mtvd$, and $\ttd$ datasets with 224x224 image observations instead of the 84x84 ones.

One combination, \emph{\appr+R3M, BC} in \Cref{fig:mw}, learns a single-task policy on 10 full sensorimotor demonstrations for each \mw\ target task. It operates in behavior cloning (BC) mode, whereby \appr\ is optimized solely w.r.t. its action predictions' MSE loss, whose gradients backpropagate though the whole network (except the frozen R3M). The other combination, \emph{pretr. \appr+R3M} in \Cref{fig:mw}, follows the same \appr\ pretraining and finetuning process as described previously, except the R3M encoder stays frozen throughout.

\textbf{Learning from Play~\cite{mcil21rss}.} Our final baseline is an adapted \emph{Learning from Play} (LfP) approach~\cite{mcil21rss}. As in \citet{mcil21rss}, LfP doesn't use video-only $\mtvd$ data or target-task demonstrations $\ttd$; it trains one model for all target tasks from the ``play'' dataset $\vmt$ only. Instead of using language annotations to separate ``meaningful'' subsequences in $\vmt$, we give LfP the ground-truth knowledge of where trajectories sampled from different tasks begin and end. Accordingly, we don't use language during training either. As n the case of \appr, We train \emph{Learning from Play} to plan conditioned only on goal images and present it with goal images from successful trajectories of the target tasks during evaluation.

\subsection{Proprioceptive states in Meta-World pretraining}

While \appr\ in general takes proprioceptive states as input (\Cref{fig:transformer}), we found that in the case of Meta-World, using them actually hurts \appr's performance. Our hypothesis is that they serve as salient features that ``distract" the $\mathcal{L}_{EX}$ loss from training the visual observation encoder and hence inducing a good latent space. The latent space, in turn, is crucial for the subsequent pretraining of \appr's planner.

Therefore, we mask out the proprioceptive states when pretraining \appr\ for Meta-World. Doing so forces the executor to learn to use the features of visual observations, not proprioceptive information, for predicting actions, shaping a better latent visual observation space for the planner.

\subsection{Success rate evaluation protocol \label[appendix]{app:srate_details}}

\textbf{In the generalization experiments on \mw}, all success rate evaluations are done on 50 500-step rollouts starting from initial states sampled from the \emph{test} distributions of \mw's ML\replaced{50}{45} target tasks (\emph{door-lock}, \emph{door-unlock}, \emph{hand-insert},  \emph{bin-picking}, and  \emph{box-close}).

To evaluate the zero-shot success rate of the pretrained EX and \appr\ models, we compute the average across 50 rollouts generated by these models on each of the 5 target tasks \emph{at the end of pretraining}.

To evaluate the success rate of the finetuned models, we adopt the procedure from \citet{robomimic2021}. The finetuning lasts for $N$ epochs (see \Cref{tab:training_params}). After each epoch, we measure the average success rate of the resulting model across 50 rollouts, and record the maximum average success rate across all finetuning epochs. 

\textbf{In the positional encoding experiments on \rsrw}, the evaluation protocol is the same as in \mw\ finetuning and in Robomimic~\cite{robomimic2021}: we train each model for $N$ epochs (see \Cref{tab:training_params}), after each epoch compute the success rate across 50 trajectories (with 700-step horizon), and record the best average success rate across all epochs.

\subsection{\rsrw\ datasets and model training \label[appendix]{app:rs_data}}

Training data for \rsrw\ was collected from human demonstrations, not generated by scripted policies. \rsrw\ provides a keyboard and SpaceMouse interfaces for controlling the Panda arm in its environments, and Robomimic supplies datasets of 200 expert (``professional-human'') trajectories collected using the SpaceMouse interface for the \emph{NutAssemblySquare}, \emph{PickPlaceCan}, and \emph{Lift} tasks. For each of the tasks without pre-collected Robomimic datasets, we gather 75 high-quality trajectories via \rsrw's keyboard interface ourselves. We employ \rsrw\ tasks only for experiments that involve training single-task policies from scratch, so all of these trajectories are used as \emph{\textbf{target-task demonstration data ($\ttd$)}}. Typical demonstration trajectory lengths vary between 50 and 300 time steps.

Accordingly, to show the difference between relative and absolute positional encodings' data efficiency, we train \appr\ for $|\ttd| =5,10,25, 50,$ and $75$, sampling $\ttd$'s from the set of 75 demonstrations without replacement. The results are presented in the main paper in \cref{fig:rsrw}.
For \emph{Lift}, \emph{PickPlaceCan}, and \emph{NutAssemblySquare}, Robomimic~\cite{robomimic2021} similarly provides 200 high-quality human-collected demonstrations each, as well as the results of BC-RNN on subsets of these datasets with $|\ttd| =40, 100,$ and $200$. Therefore, for these problems we train \appr\ for $|\ttd| =5,10,25, 50, 75$, as well as $40, 100,$ and $200$. The results are shown in \cref{tab:rs_large} and \cref{tab:plex_vs_bcrnn}.

The only difference of \appr\ model instances for \rsrw\ from those for \mw\ is the former having \emph{two} ResNet-18s in the observation encoder, one for the eye-in-hand and one for the agentview camera. As for \mw, the encoder in the  \rsrw\  is trained from scratch, in order to make our results comparable to Robomimic's~\cite{robomimic2021}, where models use an identical encoder and also train it tabula-rasa. In this experiment, we train \appr\ in behavior cloning (BC) mode, like \mw's single-task \appr+R3M, whereby \appr\ is optimized solely w.r.t. its action predictions' MSE loss, whose gradients backpropagate though the whole network. All hyperparameters are in \Cref{tab:training_params} in \Cref{app:hyp}.

We compare \appr\ with relative positional encoding to \appr\ with absolute one and to two flavors of the Decision Transformer (DT)~\cite{chen2021decision}, which use global positional embedding. One flavor (\emph{DT-global} in \Cref{fig:rsrw}) is trained to condition only on task specification (i.e., goal images), like \appr. We note, however, that \citet{chen2021decision} used rewards and returns when training and evaluating DT. Therefore, we also train a return-conditioned version of DT (\emph{DT-global(+rew)} in \Cref{fig:rsrw}), with returns uniformly sampled from the range of returns in $\ttd$ during evaluation.

\subsection{Additional \rsrw\ results \label[appendix]{app:rs}}

\textbf{Comparison to BC-RNN.} Relying on relative positional encoding allows \appr\ to achieve \emph{state-of-the art} performance on all \rsrw\ tasks in our experiments. To establish this, in addition to the baselines in  \Cref{fig:rsrw}, we compare to the results of a BC-RNN implementation from the work that introduced some of these \rsrw\ problems~\cite{robomimic2021}. Interestingly, running BC-RNN on the tasks for which we have collected demonstrations ourselves resulted in 0 success rate \replaced{}{(\Cref{tab:plex_vs_bcrnn_other})}, while running it on tasks with Robomimic-supplied 200 trajectories (\emph{Lift}, \emph{PickPlaceCan}, and \emph{NutAssemblySquare}) reproduced \citet{robomimic2021}'s results. \appr's comparison to BC-RNN's results on those problems are in \Cref{tab:plex_vs_bcrnn} in \Cref{app:rs_data}. \appr\ and BC-RNN are at par on the easier problems but \appr\ performs better on the harder \emph{NutAssemblySquare}.

\begin{table*}[h]
\setlength{\tabcolsep}{0.2em}
\begin{footnotesize}
    \centering
      \begin{tabular}{|c|c|c|c|c|c|c|c|c|c|}
        \cline{2-10}
        \multicolumn{1}{c|}{} &
          \multicolumn{3}{c|}{\emph{Lift}} &
          \multicolumn{3}{c|}{\emph{PickPlaceCan}} &
          \multicolumn{3}{c|}{\emph{NutAssemblySquare}} \\
          \cline{1-10}
        $|\ttd|$ & 40 & 100 & 200 & 40 & 100 & 200 & 40 & 100 & 200 \\
        \hline
       \appr & $100 \pm 0$ & $100 \pm 0$ & $100 \pm 0$ & $82.8 \pm 8.9$ & $95.8 \pm 2.8$ & $96.6 \pm 4.1$ & $40.4 \pm 6.9$ & $69.6 \pm 4.1$ & $86.0 \pm 3.1$ \\
        \hline
        BC-RNN & $100 \pm 0$ & $100 \pm 0$ & $100 \pm 0$ & $83.3 \pm 1.9$ & $97.3 \pm 0.9$ & $98.0 \pm 0.9$ & $29.3 \pm 4.1$ & $64.7 \pm 4.1$ & $82.0 \pm 0.0$\\
        \hline
      \end{tabular}
\caption{Performance of \appr\ and BC-RNN on three \rsrw\ tasks from ~\citet{robomimic2021} on $|\ttd|=40,100,$ and $200$ demonstrations. BC-RNN's results come from Figure 3b and Table 27 in \citet{robomimic2021}). On the easier \emph{Lift} and \emph{PickPlaceCan}, \appr\ and BC-RNN are at par, but on the harder \emph{NutAssemblySquare} \appr\ performs better. On the remaining 6 problems for which we have gathered the demonstration data, BC-RNN's success rate is 0\replaced{.}{ --- see \Cref{tab:plex_vs_bcrnn_other}.}}
\label{tab:plex_vs_bcrnn}
\end{footnotesize}  
\end{table*}

\replaced{}{\begin{table*}[h]
\setlength{\tabcolsep}{0.2em}
\begin{footnotesize}
    \centering
      \begin{tabular}{|c|c|c|c|c|c|c|}
        \cline{2-7}
        \multicolumn{1}{c|}{} &
          \multicolumn{1}{c|}{\emph{Door}} &
          \multicolumn{1}{c|}{\emph{Stack}} &
          \multicolumn{1}{c|}{\emph{PickPlaceBread}} &
          \multicolumn{1}{c|}{\emph{PickPlaceMilk}} &
          \multicolumn{1}{c|}{\emph{PickPlaceCereal}} &
          \multicolumn{1}{c|}{\emph{NutAssemblyRound}} \\
          \cline{1-7}
        $|\ttd|$ & 75 & 75 & 75 & 75 & 75 & 75  \\
        \hline
       \appr & $78.4 \pm 9.2$ & $97.3 \pm 2.9$ & $92.0 \pm 4.65$ & $65.6 \pm 4.6$ & $72.2 \pm 4.4$ & $49.8 \pm 5.5$ \\
        \hline
        BC-RNN & $0 \pm 0$ & $0 \pm 0$ & $0 \pm 0$ & $0 \pm 0$ & $0 \pm 0$ & $0 \pm 0$ \\
        \hline
      \end{tabular}
\caption{Performance of \appr\ and BC-RNN on the remaining 6 Robotsuite/Robomimic tasks from \Cref{fig:rsrw}. \appr's numbers are copied from that Figure.}
\label{tab:plex_vs_bcrnn_other}
\end{footnotesize}  
\end{table*}}

\textbf{Better data efficiency or higher performance?}  Given \Cref{fig:rsrw}, one may wonder: does \appr-abs's performance plateau at a lower level than \appr's with relative positional encoding, or does \appr-abs catch up on datasets with $|\ttd| > 75$? For most tasks we don't have enough training data to determine this, but \Cref{tab:rs_large} in \Cref{app:rs_data} provides an insight for the tasks with Robomimic-supplied 200 training demonstrations. Comparing the performance gaps between \appr\ and \appr-abs on 75-trajectory and 200-trajectory datasets reveals that the gap tends to become smaller. The same can be seen for \emph{Stack}, \emph{PickPlaceCereal}, \emph{NutAssemblyRound} already at $|\ttd| = 75$ in \Cref{fig:rsrw}, suggesting that with sufficient data \appr-abs may perform as well as \appr. However, the amount of data for which this happens may not be feasible to collect in practice.

\begin{table*}[h]
\begin{center}
  \begin{tabular}{|c|c|c|c|c|c|c|}
    \cline{2-7}
    \multicolumn{1}{c|}{} &
      \multicolumn{2}{c|}{\emph{Lift}} &
      \multicolumn{2}{c|}{\emph{PickPlaceCan}} &
      \multicolumn{2}{c|}{\emph{NutAssemblySquare}} \\
      \cline{1-7}
    $|\ttd|$ & 75  & 200 & 75 & 200  & 75 & 200 \\
    \hline
   \appr & $100 \pm 0$ & $100 \pm 0$ & $80.4 \pm 5.7$ & $96.6 \pm 4.1$ & $64.0 \pm 4.6$ & $86.0 \pm 6.1$ \\
    \hline
    \appr-abs & $100 \pm 0$ & $100 \pm 0$ & $72.8 \pm 8.0$ & $93.0 \pm 4.7$ & $45.2 \pm 5.7$ & $76.8 \pm 4.9$ \\
    \hline
  \end{tabular}
  \end{center}
  \caption{Performance of \appr\ and \appr-abs as the amount of training data $|\ttd|$ increases from 75 to 200 trajectories. The performance gap between the two is narrower on the larger dataset. For \emph{Lift} and several other \rsrw\ tasks, this trend becomes visible for datasets smaller than 200 (see \Cref{fig:rsrw}.}
  \label{tab:rs_large}
\end{table*}

\section{Hyperparameters \label[appendix]{app:hyp}}

\begin{table*}[h]
\begin{center}
  \begin{tabular}{|c|c|c|}
    \hline
    \emph{Parameter name} & \thead{\mw \\  (\emph{PLanner}/\emph{EXecutor})} & \thead{\rsrw \\ (\emph{PLanner}/\emph{EXecutor})} \\
    \hline
    \# layers & 3/3 & 3/3 \\
    context size $K$ & 30/30 time steps & 30/30 time steps \\
    hidden dimension & 256/256 & 256/256 \\
    \# transformer heads & 4/4 & 4/4 \\
    \# evaluation episodes & 50 & 50 \\
    max. evaluation episode length & 500 & 700 \\
    \hline
  \end{tabular}
  \end{center}
  \caption{Hyperparameters of \appr's transformer-based planner and executor components for the \mw\ and \rsrw\ benchmarks. In each case, the planner and executor use the same parameters, but for most problems the executor's context length $K$ can be much smaller than the planner's without loss of performance, e.g., $K_{EX}=10$. For the Decision Transformer on \rsrw, we use 4 transformer layers and otherwise the same hyperparameters as for \appr.}
  \label{tab:arch_params}
\end{table*}

\begin{table*}[h]
\begin{center}
  \begin{tabular}{|c|c|c|c|}
    \cline{2-4}
    \multicolumn{1}{c|}{} &
      \multicolumn{2}{c|}{\thead{\mw}} & \thead{\rsrw} \\
    \hline
    \emph{Parameter name} & \thead{pretraining \\  (\emph{PLanner}/\emph{EXecutor})} & \thead{last-layer finetuning \\ (\emph{PLanner}/\emph{EXecutor})} & \thead{behavior cloning \\ (\emph{PLanner}/\emph{EXecutor})} \\
    \hline
     lookahead steps & 1/ -- & 1/ -- & 1/ -- \\
    learning rate & $5\cdot 10^{-4}$ & $5\cdot 10^{-4}$ & $5\cdot 10^{-4}$ \\
    batch size & 256 & 256 & 256 \\
    weight decay & $10^{-5}$ & $10^{-5}$ & $10^{-5}$ \\
    \# training epochs & 10/10 & 10/10(?) & 10 \\
    \# training steps per epoch & 250/250 & 250/250(?) & 500 \\
    \hline
  \end{tabular}
  \end{center}
  \caption{Hyperparameters of \appr\ training for the generalization experiments on \mw\ and positional encoding experiments on \rsrw. The former use \appr\ in pretraining and finetuning modes; the latter only in behavior cloning mode (training the entire model from scratch for a single target task). In finetuning mode, we adapt only the last transformer layer of the planner and, in one experiment, of the executor as well. The (?) next to the executor's hyperparameters indicate that they were used only in the experiment where the executor was actually finetuned. For the Decision Transformer on \rsrw\, we use the same hyperparameters as for \appr.}
  \label{tab:training_params}
\end{table*}

\end{appendices}


\end{document}